\documentclass{article}

\usepackage[preprint]{neurips_2024}

\usepackage[utf8]{inputenc} 
\usepackage[T1]{fontenc}    
\usepackage{hyperref}       
\usepackage{url}            
\usepackage{booktabs}       
\usepackage{amsfonts}       
\usepackage{nicefrac}       
\usepackage{microtype}      
\usepackage{xcolor}         

\usepackage{algorithm}
\usepackage{listings}
\usepackage{graphicx}
\usepackage{amsmath}
\usepackage{amssymb}
\usepackage{booktabs}
\usepackage{multirow}
\usepackage{makecell}
\usepackage{times}
\usepackage{microtype}
\usepackage{epsfig}
\usepackage{caption}
\usepackage{float}
\usepackage{placeins}
\usepackage{color, colortbl}
\usepackage{stfloats}
\usepackage{enumitem}
\usepackage{tabularx}
\usepackage{xstring}
\usepackage{multirow}
\usepackage{xspace}
\usepackage{url}
\usepackage{subcaption}
\usepackage{siunitx}
\usepackage[british, english, american]{babel}
\usepackage{graphicx, amsmath, amssymb, caption, subcaption, multirow, overpic, textpos}

\newcolumntype{x}[1]{>{\centering\arraybackslash}p{#1pt}}
\newcolumntype{y}[1]{>{\raggedright\arraybackslash}p{#1pt}}
\newcolumntype{z}[1]{>{\raggedleft\arraybackslash}p{#1pt}}

\newcommand{\eg}{\emph{e.g.}}

\newcommand{\ie}{\emph{i.e.}}
\newcommand{\etc}{\emph{etc}}

\usepackage{pifont}  

\usepackage{wrapfig}
\usepackage{amsthm}

\usepackage{makecell}
\definecolor{baselinecolor}{gray}{.95}
\newcommand{\gr}{\rowcolor[gray]{.95}}
\definecolor{green}{HTML}{009000} 
\definecolor{red}{HTML}{ea4335}  
\definecolor{purple}{RGB}{203,15,172}
\definecolor{orange}{RGB}{235,108,21}
\newcommand{\hlg}[1]{\textcolor{green}{#1}}
\newcommand{\hlr}[1]{\textcolor{red}{#1}}

\newcommand{\worse}[1]{\hlr{$\downarrow\,$#1}}

\definecolor{00red}{RGB}{236,35,35}

\definecolor{00blue}{RGB}{50,149,237}

\definecolor{00pink}{RGB}{200,151,225}
\definecolor{02pink}{RGB}{200,151,227}

\definecolor{00grey}{RGB}{166,166,166}

\definecolor{00green}{RGB}{82,208,83}
\definecolor{02green}{RGB}{83,209,86}

\definecolor{00pink}{RGB}{230,114,138}

\definecolor{00purple}{RGB}{219,103,219}

\newcommand{\llamagrey}[1]{\cellcolor{00grey!15}{#1}}

\definecolor{F7E0D5}{RGB}{247,224,213}
\definecolor{darkF7E0D5}{RGB}{209,154,128}
\colorlet{DecPurple}{00purple!15}
\colorlet{EncBlue}{00blue!15}
\colorlet{DecRed}{00red!15}
\colorlet{DecGrey}{00grey!15}

\newlength\savewidth

\title{Adapting LLaMA Decoder to Vision Transformer}

\author{
	Jiahao Wang\textsuperscript{\rm 1},
	Wenqi Shao\textsuperscript{\rm 2*}, 
	Mengzhao Chen\textsuperscript{\rm 2}, 
	Chengyue Wu\textsuperscript{\rm 1},
	Yong Liu\textsuperscript{\rm 3},\\
        \textbf{Taiqiang Wu}\textsuperscript{\rm 1},
        \textbf{Kaipeng Zhang}\textsuperscript{\rm 2},
        \textbf{Songyang Zhang}\textsuperscript{\rm 2}, 
        \textbf{Kai Chen}\textsuperscript{\rm 2}, 
	\textbf{Ping Luo}\textsuperscript{\rm 1}\thanks{Corresponding author.} \\
	\textsuperscript{\rm 1}The University of HongKong. 
	\textsuperscript{\rm 2}Shanghai AI Laboratory. \\
	\textsuperscript{\rm 3}Tsinghua Shenzhen International Graduate School.\\
	jiahao.wang@connect.hku.hk, shaowenqi@pjlab.org.cn
}

\begin{document}

\maketitle

\begin{abstract}
This work examines whether decoder-only Transformers such as LLaMA, which were originally designed for large language models (LLMs), can be adapted to the computer vision field. 
We first "LLaMAfy" a standard ViT step-by-step to align with LLaMA's architecture, and find that directly applying a causal mask to the self-attention brings an attention collapse issue, resulting in the failure to the network training. 
We suggest to reposition the class token behind the image tokens with a \textit{post-sequence class token} technique to overcome this challenge, enabling causal self-attention to efficiently capture the entire image's information. 
Additionally, we develop a \textit{soft mask} strategy that gradually introduces a causal mask to the self-attention at the onset of training to facilitate the optimization behavior. 
The tailored model, dubbed as \textit{image LLaMA (iLLaMA)}, is akin to LLaMA in architecture and enables direct supervised learning. 
Its causal self-attention boosts computational efficiency and learns complex representation by elevating attention map ranks.
iLLaMA rivals the performance with its encoder-only counterparts, achieving 75.1\% ImageNet top-1 accuracy with only 5.7M parameters. 
Scaling the model to $\sim$310M and pre-training on ImageNet-21K further enhances the accuracy to 86.0\%. 
Extensive experiments demonstrate iLLaMA's reliable properties: shape-texture bias, calibration, quantization compatibility, ADE20K segmentation and CIFAR transfer learning. 
We hope our study can kindle fresh views to visual architectures in the wave of LLMs and inspire the development of unified multimodal models. 
Pre-trained models and codes are available \url{https://github.com/techmonsterwang/iLLaMA}. 

\end{abstract}
\section{Introduction}
\label{sec:intro}

The year 2024 saw the meteoric rise of large language models (LLMs), as well as the 4th anniversary of the Vision Transformer (ViT)~\cite{dosovitskiy2020image}. Born in 2020, ViT was influenced by the prevailing \textit{encoder-only} text Transformers at the time, such as BERT~\cite{devlin2018bert}, \etc. Accordingly, ViT is allowed to borrow the encoder-only design, \ie, self-attention do not use any causal mask. 
As a result, advanced vision backbones~\cite{touvron2021training, wang2021pyramid, guo2022cmt} and training paradigms~\cite{bao2021beit, he2022masked} have followed such convention by default. 

Meanwhile, the development of text Transformers did not stand still. A series of LLMs with a \textit{decoder-only} architecture (\eg, LLaMA~\cite{touvron2023llama}), have sparked a new wave. Pre-trained decoder-only Transformers have demonstrated remarkable performance in diverse textual~\cite{touvron2023llama2} and multimodal tasks~\cite{liu2023visual, liu2023improved, zhu2023minigpt, chen2023minigpt}. 
In this context, designing unified architectures for language and vision is a promising direction. Specifically, unified models employ shared types of operators to process both visual and textual data, reducing the cost of hardware implementation. Moreover, compared to using separate models for image and text, unified models~\cite{fuyu-8b, li2023otterhd} simplify the inference process by handling different modalities simultaneously, thereby improving practical efficiency.


\textit{Toward this goal, we took the initial step of exploring whether decoder-only Transformers can hold its effectiveness in the unimodal vision domain.} In this study, we demonstrate that through straightforward supervised learning, LLaMA architecture itself can process input images with simple yet crucial modifications. We start by modifying a standard encoder-only ViT (\eg, ViT-T/16), progressively adapting its components to align with those in LLaMA. 
In practice, we observe an \textit{attention collapse} issue, \ie, the training loss fails to converge by directly adding a causal mask to the attention map. The causal mask restricts the class token from accessing the image's global information, thereby hindering the optimization of the training loss. To this end, we propose a \textit{post-sequence class token} technique, repositioning the class token to the end of image tokens (details in Section~\ref{sec:3.3}). As a result, causal mask can keep the attention score between the class token and others, allowing the model to optimize stably. We also evaluate the advantages of the causal self-attention in reducing computational complexity and enhancing the attention map rank. 

\begin{figure*}[t]
        \vspace{-2.1em}
	\centering
	\includegraphics[width=0.92\linewidth]{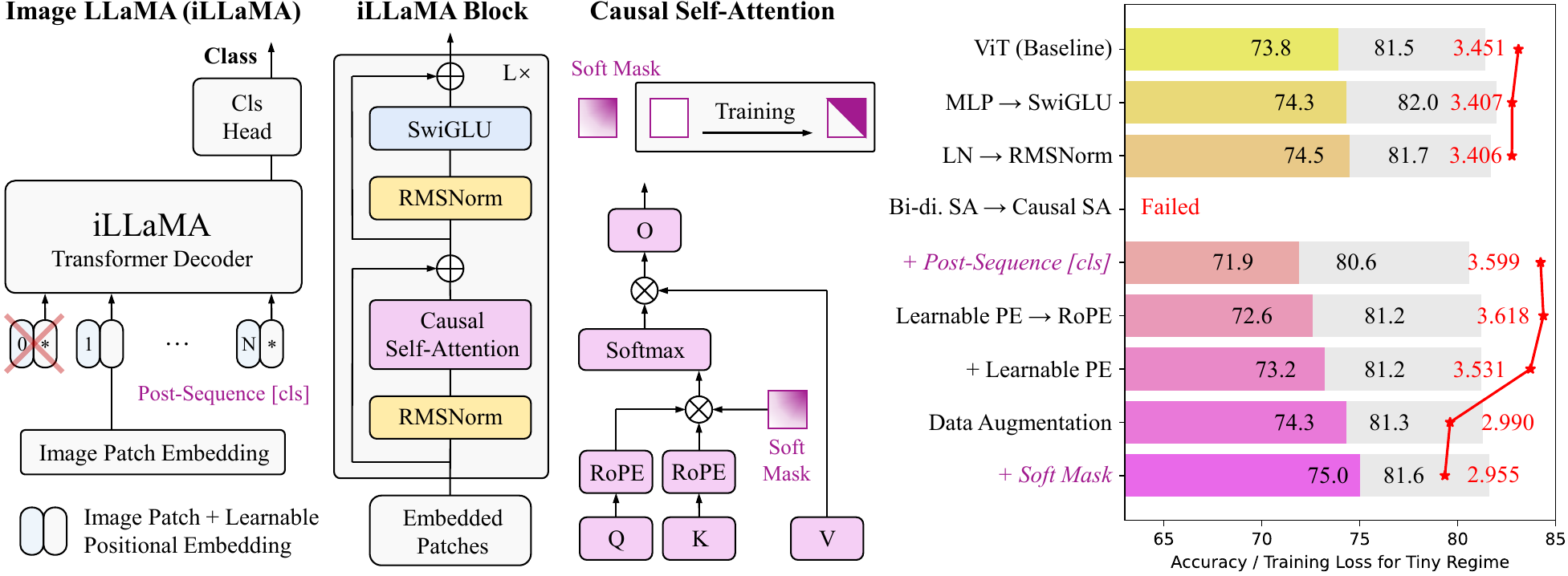}
        \vspace{-0.3em}
	\caption{\textbf{\textit{Left}: iLLaMA architecture. \textit{Right}: our design roadmap.} Colored and gray bars represent the results of the tiny and base regimes, with the red line depicting the training loss of the tiny regime. iLLaMA strives to process visual tokens using standard LLaMa components, \eg, causal self-attention. The proposed \textit{PS [cls]} and \textit{soft mask} strategy help overcome training challenges. Block details of ViT~\cite{dosovitskiy2020image}, VisionLLaMA~\cite{chu2024visionllama}, and our iLLaMA is compared in Figure~\ref{fig:comparison} in Appendix~\ref{sec:8.1}.}
        \vspace{-1.5em}
	\label{fig:architecture}
\end{figure*}

Moreover, we explore several training techniques for the proposed causal Transformer. 
When observing things, humans start by broadly catching global connections, then narrow down to focus on specifics.
Motivated by this, we develop a \textit{soft mask} approach -- bi-directional self-attention degenerates to a causal self-attention at the onset of training -- to further boost the network performance. Soft mask does not alter the causal self-attention during inference but improves the initial training behavior of the network. (details in Section~\ref{sec:3.6}). We illustrate different types of masks in Figure~\ref{fig:softmask2}. 

Equipped with these modifications, we propose a decoder-only vision Transformer with causal self-attention inside, dubbed \textit{image LLaMA (iLLaMA)}, as shown in Figure~\ref{fig:architecture}. We conduct a thorough evaluation of iLLaMA's properties, including ImageNet-1K classification~\cite{deng2009imagenet}, calibration, shape-texture bias, quantization compatibility, ADE20K semantic segmentation~\cite{zhou2019semantic}, and CIFAR transfer learning~\cite{krizhevsky2009learning}. Experimental results show that iLLaMA delivers favorable and reliable performance to its encoder-only counterparts (\ie, ViT, VisionLLaMA), while maintaining a pure decoder design. 
More importantly, a spectral analysis on the attention map empirically shows that compared to bi-directional counterparts, causal self-attention has a higher rank (see Figure~\ref{fig:rank_analysis}), which allows for learning complex image representation. The contribution of our work can be summarized as follows:

\begin{itemize}[leftmargin=*,topsep=0pt,itemsep=0pt,noitemsep]
\item We explore the adaption of the LLaMA decoder to visual tasks and identify the attention collapse issue caused by causal self-attention. To address this, we introduce the \textit{PS [CLS]} method. 
\item We propose a \textit{soft mask} strategy to optimize causal self-attention and analyzed the improvement in the rank of the causal attention map. 
\item We develop a series of \textit{iLLaMA} models and empirically validate its performance on ImageNet, along with practical properties such as quantization compatibility, calibration, shape-texture bias. 
\end{itemize}

We hope our work to inspire a re-evaluation of vision backbone design in the era of LLMs and provide fresh insights for their architectural unification.

\section{Preliminaries and Motivation}
\label{sec:prelimi}

\textbf{Encoder and decoder.} We briefly summarize the encoder and decoder in Transformer~\cite{vaswani2017attention}. Both of them basically consist of attention module and a MLP module, each followed by a residual connection. \textit{The key difference between them is the mask scheme in their self-attention}. Encoders use bi-directional self-attention, and decoders employ causal self-attention and cross-attention. However, the latter is typically omitted in decoder-only 
LLMs~\cite{touvron2023llama, touvron2023llama2}, we thus focus on comparing causal and bi-directional self-attention as follows, in terms of the \textit{mask} setting. Denote ${\bf X}\in\mathbb{R}^{N\times d}, {\bf O}\in\mathbb{R}^{N\times d}$ as the input and output sequences, where $N$ is the number of tokens and $d$ is the embedding dimension. $W_{\bf q}, W_{\bf k}, W_{\bf v}\in\mathbb{R}^{d\times d}$ denotes the linear mapping of query, key and value, respectively. Generally, self-attention module can be formulated as (set the head number and batch size as $1$ for simplicity): 
\begin{equation}
\small
{\bf A} = \frac{1}{\sqrt{d}}(W_{\bf q}({\bf X}) \cdot W_{\bf k}({\bf X})^{\top}), ~~~
{\bf O} =\operatorname{Softmax}({\bf A} + {\bf M})\cdot W_{\bf v}({\bf X}),~~~ {\bf P}_{i,j} = 0,~~~{\bf Q}_{i,j}=\left\{
\begin{aligned}
0 & , i\ge j \\
-\infty & , i<j 
\end{aligned}
\right.
\label{eq:preliminary1}
\end{equation}
where $i,j\in [1,N]$, ${\bf A}\in\mathbb{R}^{N\times N}$, ${\bf M}\in\mathbb{R}^{N\times N}$ denote the attention map and mask. ${\bf P}\in\mathbb{R}^{N\times N}$, ${\bf Q}\in\mathbb{R}^{N\times N}$ are masks in the encoder and decoder, respectively. 
For a causal self-attention, we have ${\bf M}={\bf Q}$. 
Such design allows subsequent tokens only attend to the preceding ones, but not vice versa. For a bi-directional self-attention, we have ${\bf M}={\bf P}$, ensuring mutual 
visibility for each token.



\textbf{Recent LLMs-related image models.} Recent image models~\cite{bai2023sequential, guo2024data, el2024scalable} are trained with an autoregressive objective, targeting at solving visual tasks. Pang et al.~\cite{pang2023frozen} add a text pre-trained frozen LLM block to a ViT encoder to facilitate the performance. Our work, on the other hand, is motivated to explore in-depth how the decoder design in LLMs can be adapted to image models using simple supervised learning to achieve an architectural alignment. A concurrent work VisionLLaMA~\cite{chu2024visionllama} proposes vision models for recognition and generation tasks based on the LLaMA components. Differently, we: 1) introduce causal self-attention from LLaMA, addressing the associated attention collapse issue, while VisionLLaMA retains an encoder architecture; 2) develop a soft mask technique to assist training the decoder; 3) expand the dataset to the larger ImageNet-21K to demonstrate scalability, achieving 86.0\% ImageNet accuracy that outperforms VisionLLaMA's best results.

\section{A Roadmap: Solving Attention Collapse and Optimization Improvement}
\label{sec:method_vision}

This section introduces the design roadmap of iLLaMA. As we aim to adapt LLMs to vision, we choose LLaMA~\cite{touvron2023llama} and ViT~\cite{dosovitskiy2020image} as language and vision baselines due to their successful practices.
The trajectory can be divided into two dimensions, \ie, architecture (Section~\ref{sec:3.1}-\ref{sec:3.4}) and training techniques (Section~\ref{sec:3.5}-\ref{sec:3.6}). First, we focus on block designs including 1) feed foward network, 2) normalization layer, 3) self-attention, 4) positional embedding, illustrated in Figure~\ref{fig:architecture}. Next, we study training techniques and develop a soft mask strategy to facilitate optimization. 
Finally, we provide an analysis in terms of efficiency and attention map rank (Section~\ref{sec:3.7}). 
We start with ViT-T/16 and ViT-B/16 with around 5.7M and 86.4M parameters, respectively, and gradually replace the corresponding components with those from LLaMA. We conduct experiments on ImageNet-1K~\cite{deng2009imagenet}, following the training recipe adopted from~\cite{liu2023dropout} (details in Table~\ref{tab:setup_basic} of Appendix~\ref{subsec:8.3.1}). Considering the differences between visual perception and text generation tasks, we maintain ViT's non-autoregressive manner in our network. Each step change and the corresponding results are reported in Table~\ref{tab:exploration} of Appendix~\ref{sec:8.4}.

\subsection{Feed Forward Network (FFN)} 
\label{sec:3.1}

FFN in Transformer are implemented differently in ViT and LLaMa, \ie, multi-layer perceptron (MLP) and SwiGLU~\cite{shazeer2020glu}. MLP consists of two sequential linear mappings, with a GELU~\cite{hendrycks2016gaussian} function inserted. Meanwhile, SwiGLU combines three linear mappings, integrating a SiLU~\cite{hendrycks2016gaussian, elfwing2018sigmoid, ramachandran2017searching} function. This structure allows for the modulation of high-dimensional features through a gating mechanism.
We substitute the Transformer's MLPs with SwiGLUs, while maintaining comparable computational cost. As shown in Figure~\ref{fig:architecture}, this improves performance from $73.8\%$ to $74.3\%$, and from $81.3\%$ to $82.0\%$ for the ViT-T/16 and ViT-B/16 regime.
This highlights SwiGLU's effectiveness not only in language models but also in vision, inspiring further exploration of other components. 

\textit{We will now use SwiGLU to substitute MLP in each block.}

\subsection{Normalization Layer} 
\label{sec:3.2}
Transformers need normalization layer for stable training, \ie, layer normalization (LN)~\cite{ba2016layer} in ViT and root mean square layer normalization (RMSNorm)~\cite{zhang2019root} in LLaMA, respectively. We replace all LNs with RMSNorms in our network and empirically observed that the accuracy of the ViT-T/16 regime increased from $74.3\%$ to $74.5\%$. However, similar improvements in precision were not observed in the ViT-B/16 regime (from $82.0\%$ to $81.7\%$). Nonetheless, compared to LN, RMSNorm removes the shift term computation, bringing simplicity to the network~\cite{touvron2023llama2, vicuna2023, roziere2023code, jiang2023mistral}. 

\textit{We will use RMSNorm instead of LN as the normalization layer in each block.}

\begin{figure*}[t]
\vspace{-1.1em}
\centering
\begin{tabular}{ccc}
\includegraphics[width=0.174\linewidth]{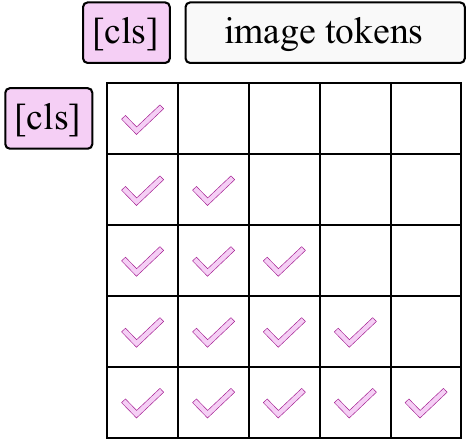} &~~~~~~~~~~~~~~
\includegraphics[width=0.18\linewidth]{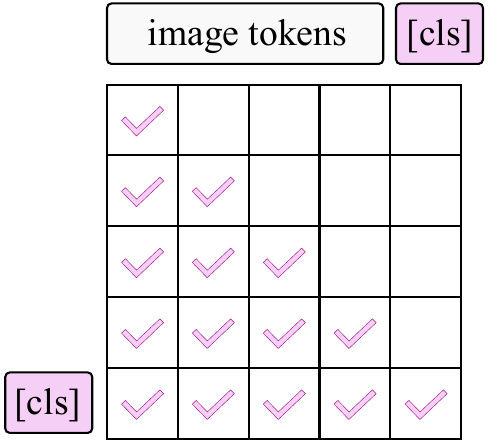} &
\includegraphics[width=0.17\linewidth]{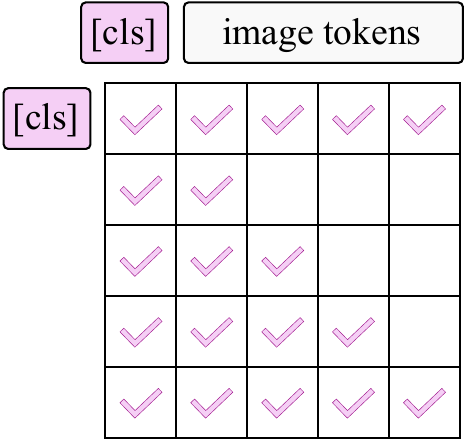}~~ \\
~(a) causal mask & ~~~~ (b) causal mask w/ \textit{PS [cls]} (ours) & ~~ (c) modified causal mask (ablation) \\
\end{tabular}
\vspace{-0.2em}
\caption{(a) mask in causal self-attention. (b) mask in causal self-attention with our post-sequence class token (\textit{PS [cls]}) method. (c) modified causal mask. Their ablation results are shown in Table~\ref{tab:abl_post}. 
}
\label{fig:mask3}
\vspace{-1.1em}
\end{figure*}

\subsection{Causal Self-Attention Leads to Attention Collapse} 
\label{sec:3.3}
\textbf{Attention collapse issue.} As a common practice for Transformer decoders, the key component for causal self-attention is the causal mask, \ie, a lower triangular mask matrix, illustrated in Eq.~\ref{eq:preliminary1} and Figure~\ref{fig:mask3}(a). With such, each token can get the attention score of all its previous ones.
We add the causal mask to our network via a non-autoregressive way. 
The reason is that visual perception tasks, unlike text generation, require only inference once. 
As a result, we observe that the training loss fails to converge in both ViT-T/16 and ViT-B/16 regimes (line 1 in Table~\ref{tab:abl_post}). We posit that such issue stems from the influence of the lower triangular matrix, which prevents the class token from "seeing" other image tokens. As illustrated in Figure~\ref{fig:mask3}(a), when the class token is positioned at the start of the patch embedding, its attention score for all other image tokens gets zero due to a causal mask. We term this occurrence as the \textit{attention collapse} issue, which leads to a loss of connection between the class token and other image patches, thereby hindering the optimization of the network. 

\begin{wraptable}{r}{0.5\textwidth}
    \centering
    \vspace{-1.3em}
    \small
    \caption{Results of \textit{PS [cls]} and the modified causal mask. Training converges in both settings.}
    \vspace{-0.5em}
    \label{tab:abl_post}
    \renewcommand{\arraystretch}{1.15}
    \setlength{\tabcolsep}{5.5pt}
    \begin{tabular}{l|cccc}
        \toprule[1.5pt]
        Model & Tiny & Train Loss & Base & Train Loss \\
        \midrule
        None & 0.1 & Failed & 0.1 & Failed \\
        \gr PS [cls] & 71.9 & 3.599 & 80.6 & 2.869 \\
        Modified & 72.5 & 3.550 & 80.4 & 2.857 \\ 
        \bottomrule[1pt]
    \end{tabular}
    \vspace{-1em}
\end{wraptable} 

\textbf{Post-sequence class token (\textit{PS [cls]}).} 
The attention collapse issue stems from the inappropriate placement of the token. To this end, we suggest a \textit{PS [cls]} strategy, by placing it at the end of the token sequence, without changing the causal mask, as shown in Figure~\ref{fig:mask3}(b) and Figure~\ref{fig:architecture}. Such modification ensures that the class token can achieve global information about all image tokens, while maintaining a causal self-attention property. 
As a result, we observe that the attention collapse issue is eliminated and the training process starts to stabilize, leading the network performance to $71.9\%$ for ViT-T/16 and $80.6\%$ for ViT-B/16 regime, respectively (line 2 in Table~\ref{tab:abl_post}). 

To test our hypothesis about the reason of the attention collapse issue, we also explore a mask setting in Figure~\ref{fig:mask3}(c). In this setting, we do not change the position of the class token. Instead, we unmask the first row of the mask (\ie, attention score of the class token) on the basis of the causal self-attention, termed as "modified causal mask". Ablation results (line 3 in Table~\ref{tab:abl_post}) shows that both settings can solve the attention collapse issue as expected, and the "modified causal mask" leads to a better $72.5\%$ accuracy for ViT-T/16 regime, validating our hypothesis about the reason. 
Although the results do not surpass the performance of bi-directional counterpart, they demonstrate the potential for optimizing causal self-attention in a decoder-only image model. 
We also observe that the \textit{PS [cls]} method yields higher accuracy with a slightly larger training loss for ViT-B/16 regime, suggesting lower overfitting. 

\textit{We will employ causal self-attention with \textbf{the proposed \textit{PS [cls]} method} in each block.}

\subsection{Positional Embedding} 
\label{sec:3.4}
A standard ViT use learnable positional embedding (LPE) to preserve positional information, typically adding it directly to the patch embedding. Meanwhile, rotary positional embedding (RoPE)~\cite{su2024roformer} is widely employed in LLMs~\cite{touvron2023llama, touvron2023llama2}, which functions in the attention of each block. 
We first use RoPE alone, which boosts the accuracy of ViT-T/16 and ViT-B/16 regimes to $72.6\%$ and $81.2\%$, from $71.9\%$ and $80.6\%$, respectively. 
The encouraging results illustrate that the concepts of "position" in image and text do not exist an inherent gap. Since LPE functions only once before all Transformer blocks, keeping it does not disrupt the alignment with LLaMA within each block. Thus, we reintroduce the LPE, which improves the accuracy of ViT-T/16 regime from $72.6\%$ to $73.2\%$, suggesting that the two positional embeddings are not redundant but rather synergistic, enhancing network performance. 

\textit{We will use both LPE and RoPE for positional embedding. So far, we have investigated each block component, and thus fix the final architecture dubbed iLLaMA. Next, we explore training strategies.}

\subsection{Data Augmentation} 
\label{sec:3.5}
Mixup~\cite{zhang2017mixup} and cutmix~\cite{yun2019cutmix} that we used to train our iLLaMA (0.8 and 1.0), are borrowed from DeiT~\cite{touvron2021training}'s recipe. Unlike the bi-directional self-attention used in DeiT, causal self-attention affects the connection between image tokens. Meanwhile, these two hyper-parameters affect the content of the input image, which further influences the subsequent embedding. Thus, we reevaluate their impact on iLLaMA optimization. Specifically, we discover that a combination of 0.1 mixup and 0.1 cutmix improves the performance of the iLLaMA-T/16 from $73.2\%$ to $74.3\%$, whereas a combination of 0.95 and 1.0 leads the iLLaMA-B/16 to a $81.3\%$ accuracy. Other ablations are detailed in Section~\ref{sec:4.1}.

\subsection{Soft Mask Strategy: Optimization Improvement} 
\label{sec:3.6}
When observing objects, humans tend to perceive broad connections, then focus on specific details. Motivated by this, we propose a \textit{soft mask} technique to improve the model's optimization: \textit{starting with bi-directional self-attentions in the early training epochs and gradually shifting completely to causal self-attentions as the optimization goes}. Specifically, self-attention can be formulated as: 

\begin{equation}
\small
\begin{aligned}
{\bf A} &= \frac{1}{\sqrt{d}}(W_{\bf q}({\bf X}) \cdot W_{\bf k}({\bf X})^{\top}), ~~~{\bf O} =(\operatorname{Softmax}({\bf A})\odot {\bf S})\cdot W_{\bf v}({\bf X}),
\\ 
{\bf S} &=\alpha {\bf B} +(1-\alpha){\bf C}, ~~~
{\bf B}_{i,j} = 1,~~~
{\bf C}_{i,j}=\left\{
\begin{aligned}
1 & , i\ge j \\
0 & , i<j 
\end{aligned}
\right.
\end{aligned}
\label{eq:softmask1}
\end{equation}

\begin{figure*}[t]
\vspace{-1.1em}
\centering
\begin{tabular}{cc}
\includegraphics[width=0.47\linewidth]{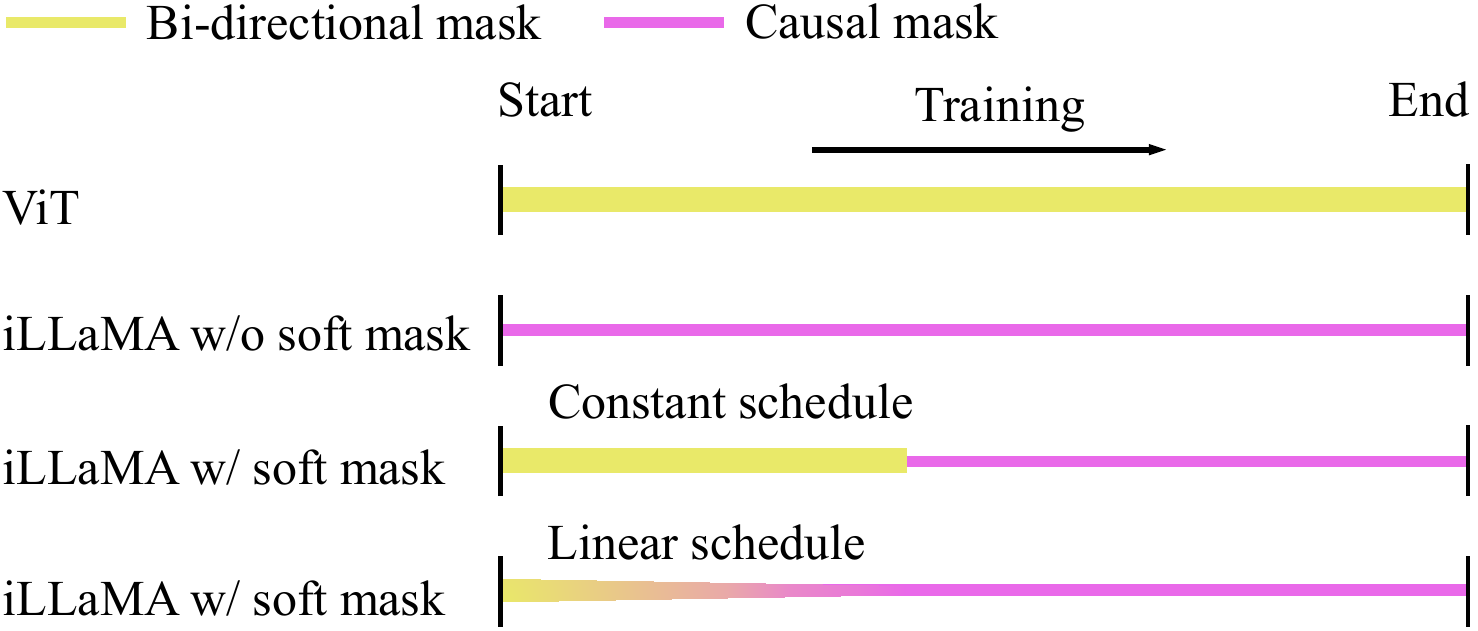} &
\includegraphics[width=0.4\linewidth]{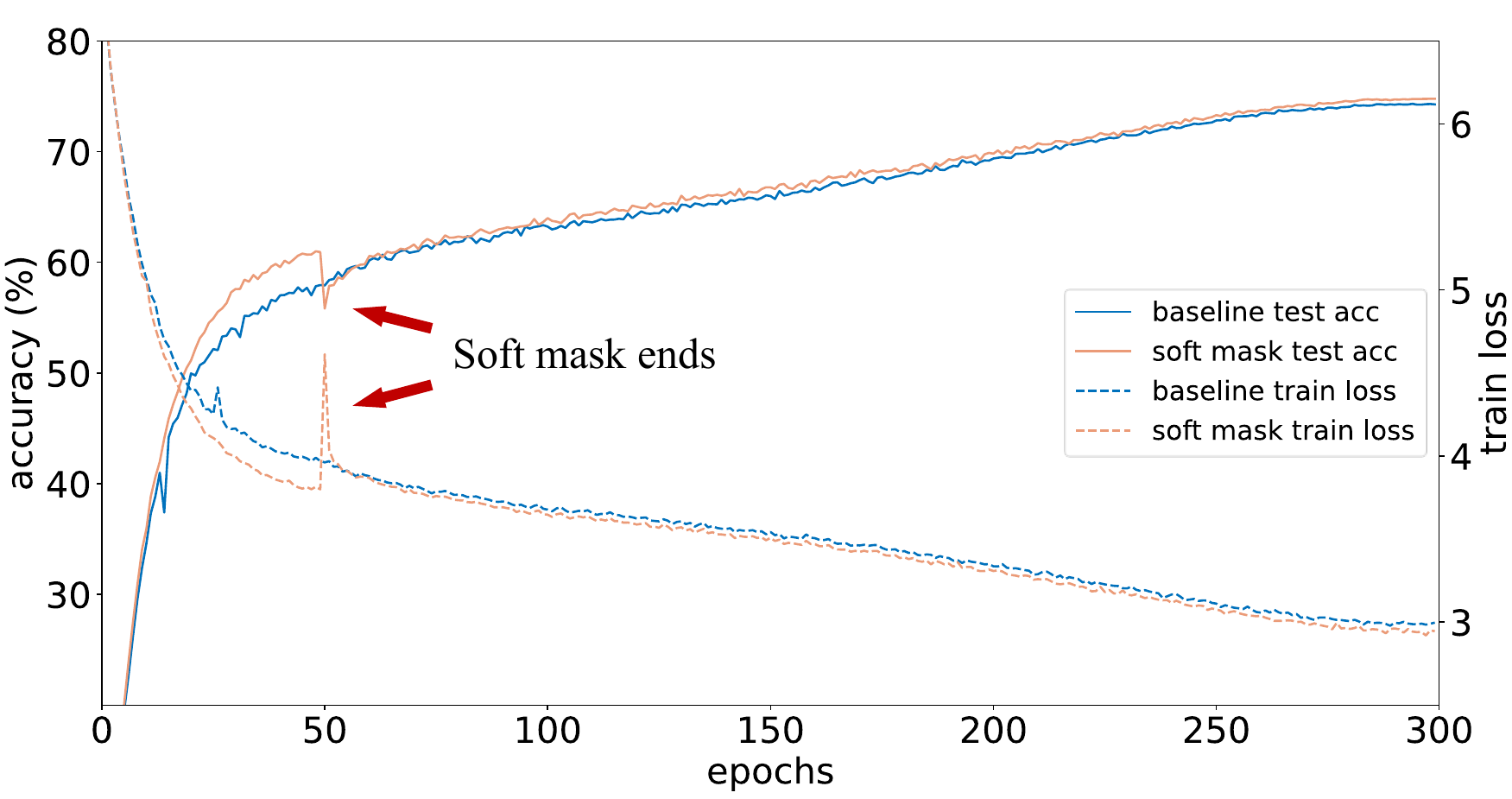} \\
~~~(a) soft mask scheme & (b) training curves w/ or w/o soft mask \\
\end{tabular}
\vspace{-0.2em}
\caption{(a) Soft mask gradually transitions from a bi-directional mask into a causal mask during training through a constant or linear schedule. (b) Ablation results of training loss and test accuracy.}
\label{fig:softmask2}
\vspace{-1.4em}
\end{figure*}
where $i,j \in [1,N]$, ${\bf S}\in\mathbb{R}^{N\times N}$ denotes the soft mask, which is defined as a linear combination of a bi-directional mask ${\bf B}$ and a causal mask ${\bf C}$. $\alpha$ is the hyper-parameter controlling the mask configuration, \ie, soft mask degenerates into ${\bf B}$ or ${\bf C}$ when $\alpha=1$ or $\alpha=0$, respectively. As illustrated in Figure~\ref{fig:softmask2}(a), $\alpha$ involves three related hyper-parameters: 1) scheme: how $\alpha$ drops from 1 to 0: we try a linear or a constant scheme. 2) cutoff epochs: when will $\alpha$ drops to 0. 3) learning rate (lr) warmup~\cite{he2016deep, goyal2017accurate}: this hyper-parameter overlaps with the duration of soft mask. We initially set the lr warmup epochs at 50, consistent with previous settings. When using a linear scheme with 50 and 25 cutoff epochs, we observe an improvement in performance for both iLLaMA-T/16 and iLLaMA-B/16 models, reaching $74.9\%$ and $81.6\%$ from $74.3\%$ and $81.3\%$, respectively. Ablations results are detailed in Section.~\ref{sec:4.1}. 
To intuitively observe the impact of soft mask, we plot the training curve of the iLLaMA-T/16 in Figure~\ref{fig:softmask2}(b), using a constant scheme with 50 cutoff epochs. When soft mask ends, we observe that although there was a sharp drop in accuracy, the model ends up achieving better performance. Similar case of the iLLaMA-B/16 are shown in Figure~\ref{fig:train_loss_base} of Appendix~\ref{sec:8.6}. Additionally, we discover that a lower learning rate warmup helps iLLaMA-T/16 achieve $75.0\%$ top-1 accuracy, by using a constant scheme with 50 cutoff epochs.  Therefore, we use this warmup method for iLLaMA-T/16. Notably, the final training loss within both iLLaMA-T/16 and iLLaMA-B/16 decreases when using soft masks, suggesting a mitigation of potential underfitting concerns.

\subsection{Analysis of causal Self-attention}
\label{sec:3.7}

Next, we analyze the advantages of using causal self-attention in iLLaMA, in terms of computational efficiency and expressive ability of visual representation through the lens of attention map rank. 

\textbf{Computational Complexity.} We compare the efficiency of causal self-attention and bi-directional baseline. For a self-attention with a sequence length $N$ and embedding dimension $D$, FLOPs are reported in Table~\ref{tab:flops} (RoPE is not involved as only attention computations are considered). causal self-attention, in light of the lower triangular property of its attention map, slightly reduces the FLOPs compared to the bi-directional baseline --- the degree of reduction grows as sequence length increases. 

\begin{wraptable}{r}{0.58\textwidth}
    \vspace{-1em}
    \centering
    \small
    \vspace{-0.2em}
    \caption{Computational complexity results. causal mask slightly reduces FLOPs required in the self-attention.}
    \vspace{-0.7em}
    \label{tab:flops}
    \renewcommand{\arraystretch}{1.13}
    \setlength{\tabcolsep}{2.2pt}
    \begin{tabular}{l|c|c}
        \toprule[1.5pt]
        Type & Bi-directional & causal \\
        \midrule
        FLOPs & $4ND^2 + 2N^2D$ & $4ND^2 + N^2D + (\lfloor N^2/2 \rfloor+1) D$ \\ 
        \bottomrule[1pt]
    \end{tabular}
    \vspace{-1.1em}
\end{wraptable}

\begin{wrapfigure}{r}{0.44\textwidth}
\vspace{-1.1em}
\centering
\small
\includegraphics[width=1.0\linewidth]{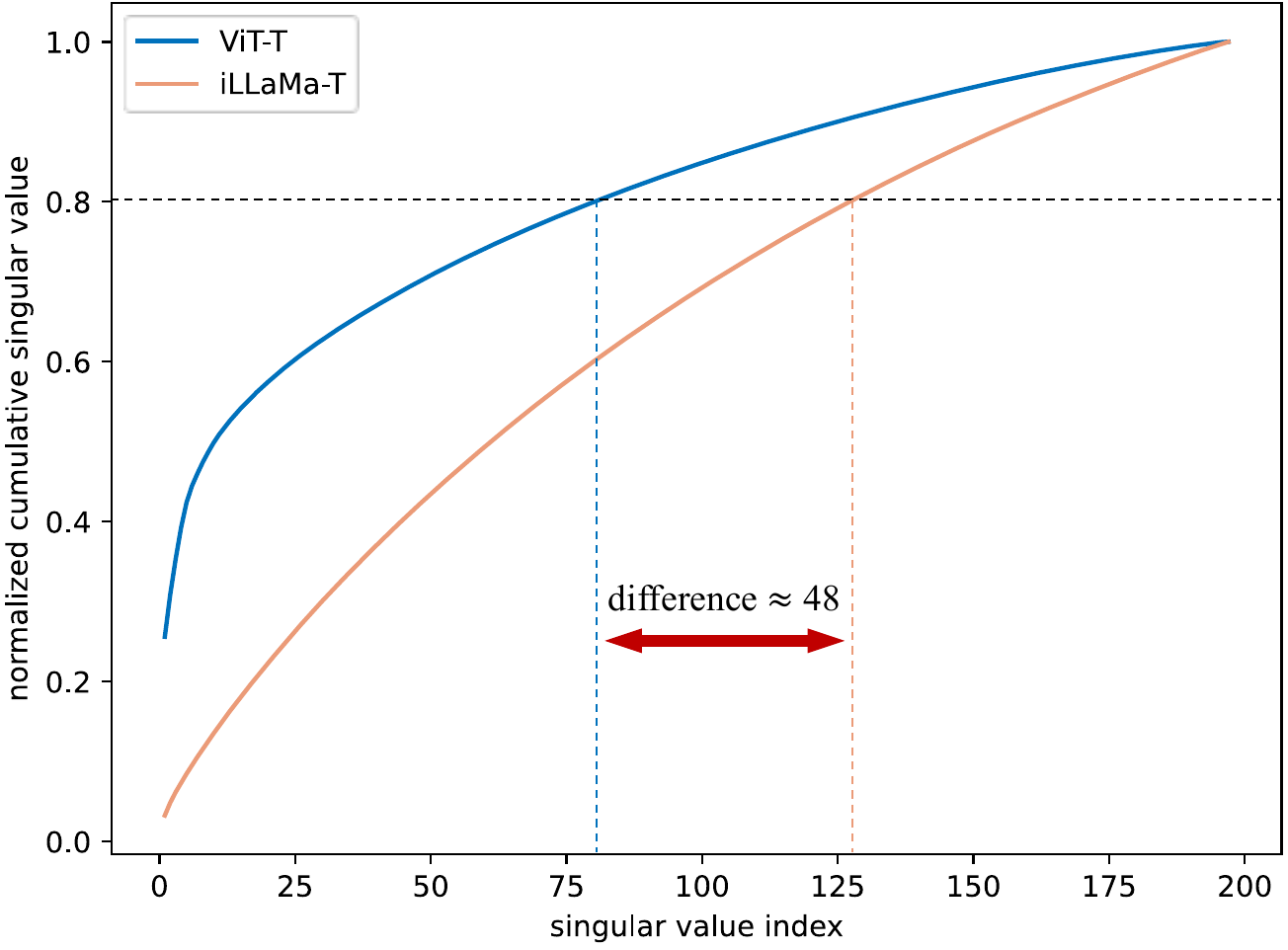}
\vspace{-1.2em}
\caption{Rank analysis of the attention map in head 1, layer 1 of the pretrained ViT-T and iLLaMA-T with $N=197$. Difference between them is about 48.}
\label{fig:rank_analysis}
\vspace{-1.6em}
\end{wrapfigure}

\textbf{Attention map rank.} We examine the representation learning power of causal attention through a spectrum analysis. Following~\cite{wang2020linformer, shu2021adder}, we perform singular value decomposition on the attention maps of the pre-trained ViT-T/16 and iLLaMA-T/16 models.
Next, we sort the singular values and plot a curve illustrating the relationship between the cumulative normalized singular values and matrix indices. The results are conducted using $30$ images randomly selected from the ImageNet-1K validation set. As shown in Figure~\ref{fig:rank_analysis}, the curve of ViT exhibits concave function characteristics, while the curve of iLLaMA is close to a linear function, indicating a more uniform distribution of singular values in iLLaMA's attention map. Approximating the matrix rank by the index at which the cumulative normalized singular value reaches 0.8, we observe that the index value of iLLaMA is about 48 higher than that of ViT ($\sim$129-th v.s. $\sim$81-th). Under such premise, compared to ViT, the attention map of iLLaMA can be approximated with a certain error by a higher-rank matrix. Accordingly, the rank of the attention map may affect the expressive capabilities of the learned representations~\cite{dong2021attention}, suggesting that the causal self-attention in iLLaMA has the potential to learn complex visual representations, as empirically demonstrated in Section~\ref{sec:4.2}. Detailed results for different layers are provided in Figure~\ref{fig:rank12} of Appendix~\ref{sec:8.5}. 

\textbf{Closing remarks.} 
So far, we have finished the design roadmap of iLLaMA through architectural and training strategy modification.
iLLaMA, a decoder-only Transformer, shows advantages in computational complexity and attention map rank through its causal self-attention mechanism. Notably, while all components of iLLaMA are essentially derived from LLaMA, relying only on them is insufficient for an effective weight optimization, as demonstrated in Section~\ref{sec:3.3}. In fact, the proposed \textit{PS [cls]} and soft mask strategy effectively address this issue and assist in iLLaMA training. However, to achieve a comprehensive understanding of iLLaMA's properties, some useful evaluation should be conducted:
1) Scalability for large model  capacities (>300M parameters) and dataset sizes (>10M training images, \eg, ImageNet-21K). 2) Other practical evaluation dimensions, such as model calibration, shape-texture bias, downstream task performance, quantization compatibility, discussed below. 
\section{Experiments}
\label{sec:experiments}
This section provide a comprehensive evaluation of iLLaMA. We first report ablation results, \eg, the effectiveness of data augmentation and different soft mask strategies. Next, we compare iLLaMA with other strong baselines on ImageNet classification. Beyond ImageNet accuracy, we also examine its efficacy on calibration, shape-texture bias, and evaluate its compatibility with quantization-aware training and downstream task performance. 

\subsection{Ablation Study} 
\label{sec:4.1} 

\textbf{Influence of data augmentation.} Base on the observation in Section~\ref{sec:3.5}, we examined multiple sets of cutmix and mixup settings, as reported in Table~\ref{tab:da_schedule}. We empirically observe that the smaller iLLaMA-T/16 are more sensitive to two data augmentation strategies and perform better with lower hyper-parameters, whereas the larger iLLaMA-B/16 are suited to higher ones. This may be related to the architectural differences between LLaMA's Transformer decoder and ViT's encoder type. 

\textbf{Influence of soft mask scheduling strategies and epochs.} As mentioned in Section~\ref{sec:3.6}, the proposed soft mask technique includes three hyper-parameters, \ie, schedule, cutoff epochs and lr warmup epochs. Here we evaluate the robustness of soft mask to hyper-parameter settings, with results detailed in Table~\ref{tab:abl_schedule}. Beyond the \textit{linear} schedule, inspired by~\cite{liu2023dropout}, we also implemented a \textit{constant} option. Additionally, we fixed the learning rate warm-up epochs at 50 and experimented with different cutoff epochs. The results reveal that the soft mask facilitates the optimization of iLLaMA under both linear and constant scheduling, suitable for models of both tiny and base sizes. Moreover, setting the cutoff epochs to span a wide range from 25 to 100 is advantageous. Notably, the soft mask can be easily integrated into existing code frameworks (\eg, timm~\cite{pytorchmodels}) with negligible additional training costs, thereby facilitating its effortless application on future related architectures. 

\begin{table}[t]
        \vspace{-1.2em}
	\begin{minipage}[t]{0.44\linewidth}
		\centering
        \small
		\renewcommand{\arraystretch}{1.05} 
		\caption{Soft mask scheduling for iLLaMA-T/16 and iLLaMA-B/16 on ImageNet-1K.}\label{tab:abl_schedule}
		\vspace{0.2em}
		\setlength{\tabcolsep}{6pt}
            \begin{tabular}{lccc}
                \toprule[1.5pt]
                Schedule & Cutoff Epochs & Tiny & Base \\
                \midrule
                no softmask  & - & 74.3 & 81.3 \\
                \midrule
                \gr linear & 25 & 74.8 & \textbf{81.6} \\
                \gr linear & 50 & \textbf{74.9} & 81.5 \\
                linear & 100 & \textbf{74.9} & 81.5 \\
                constant & 25 & 74.7 & 81.5 \\
                constant & 50 & 74.8 & 81.5 \\
                \bottomrule[1pt]
            \end{tabular}
	\end{minipage}
	\hspace{.15in}
	\begin{minipage}[t]{0.52\linewidth}
		\centering
        \small
		\renewcommand{\arraystretch}{1.13} 
		\caption{Soft mask for training loss and testing loss. Soft mask lowers both training and testing loss in tiny and base models, counteracting underfitting issue and thus leading to a better optimization.}\label{tab:abl_softmask}
		\vspace{0.2em}
		\setlength{\tabcolsep}{9pt}
            \begin{tabular}{lcc}
            \toprule[1.5pt]
            Model & Training Loss & Testing Loss \\
            \midrule
            tiny & 2.990 & 1.121 \\
            \gr + soft mask &  2.955 (\worse{0.045}) & 1.092 (\worse{0.029}) \\
            base & 2.868 & 0.843 \\
            \gr + soft mask &  2.828 (\worse{0.040}) & 0.831 (\worse{0.012}) \\
            \bottomrule[1pt]
            \end{tabular}
	\end{minipage}
	\vspace{-1.8em}
\end{table}


\begin{wraptable}{r}{0.44\textwidth}
    \vspace{-1.2em}
    \centering
    \small
    \caption{Mixup and cutmix ablation results.}
    \label{tab:da_schedule}
    \renewcommand{\arraystretch}{1.05}
    \vspace{-0.6em}
    \setlength{\tabcolsep}{3pt}
    \begin{tabular}{ccc|ccc}
        \toprule[1.5pt]
        Mixup & Cutmix & Tiny & Mixup & Cutmix & Base \\
        \midrule
        0.8  & 1.0 & 73.2 & 0.8  & 1.0 & 81.2 \\
        0.5  & 0.4 & 73.8 & 0.9  & 0.9 & 81.2 \\
        0.3  & 0.3 & 73.9 & 0.9  & 1.0 & 81.2 \\
        0.2  & 0.2 & \textbf{74.3} & 1.0  & 1.0 & 81.2 \\
        \gr 0.1 & 0.1 & \textbf{74.3} & 0.95  & 1.0 & \textbf{81.3} \\
        \bottomrule[1pt]
    \end{tabular}
    \vspace{-1.2em}
\end{wraptable}


\textbf{Influence of soft mask for training and testing loss.} 
Deep neural networks often face underfitting, marked by difficulty in continuously reducing training loss and resulting in poor test accuracy~\cite{liu2023dropout}. 
We compare the training and testing losses of the iLLaMA-T/16 and iLLaMA-B/16 models with and without the use of the soft mask strategy. As shown in Table~\ref{tab:abl_softmask}, soft mask can reduce training loss in both regimes, mitigating potential underfitting issue and reducing testing loss.

\subsection{Comparison with Recent Architectures on ImageNet-1K Classification} 
\label{sec:4.2}

We conducted experiments on the ImageNet-1K~\cite{deng2009imagenet} benchmark with different model sizes (\ie, iLLaMA-T/S/B/L). Detailed architecture configurations are shown in Table~\ref{tab:hyper} of Appendix~\ref{sec:8.1}. Our ImageNet-1K/21K (pre-)training and ImageNet-1K fine-tuning recipes are shown in Table~\ref{tab:setup_pretrain} and Table~\ref{tab:setup_finetune} of Appendix~\ref{sec:8.3}. 
We also study the use of LLaMA2-7B pre-trained weights for iLLaMA initialization, and the paradigm and results are detailed in Figure~\ref{fig:weight_selection} and Table~\ref{tab:weight_selection} in Appendix~\ref{sec:8.9}. 

\textbf{ImageNet-1K training.} We train iLLaMA-T/S/B on ImageNet-1K for 300 epochs with AdamW optimizer~\cite{loshchilov2017decoupled} and a batch size of 4096. The ImageNet-1K trained iLLaMA-T/B models are, in fact, the outcome of the explorations completed in Section~\ref{sec:3.6}. For the settings of soft mask schedule, cutoff epochs, and learning rate warmup epochs, we tune slightly for the iLLaMA-S model. 

\textbf{ImageNet-21K pre-training.} We use the `Winter21 variant of ImageNet-21K-P' (refered to as ImageNet-21K) dataset~\cite{ridnik2021imagenet} \footnote{downloaded from: https://www.image-net.org/download-images.php} for the large-scale pre-training of our iLLaMA, which contains 11,060,223 training images and 522,500 testing images from 10,450 classes. Only the training set was used. We pre-train iLLaMA-B/L on ImageNet-21K for 90 epochs using a constant soft mask schedule, with cutoff epochs and learning rate warmup epochs set to 30 and 5, respectively. 

\textbf{ImageNet-1K fine-tuning.} For iLLaMA-B model trained on ImageNet-1K, we fine-tune at a resolution of 384$\times$384. Similarly, for the iLLaMA-B/L model trained on ImageNet-21K, we fine-tune at resolutions of 224$\times$224 and 384$\times$384, respectively. All fine-tuning was conducted for 30 epochs using the AdamW optimizer. We follow DeiT~\cite{touvron2021training} for interpolating positional embeddings to allow our iLLaMA to handle inputs at a higher resolution.  

\textbf{Results.} Table~\ref{tab:imagenet} shows a comparison between iLLaMA and other strong visual baselines, including ConvNets (ConvNeXt~\cite{liu2022convnet}, ConvNeXt-V2~\cite{woo2023convnext}), vision Transformers (ViT~\cite{dosovitskiy2020image}, Swin Transformer~\cite{liu2021swin}), MLPs (PoolFormer~\cite{yu2022metaformer}, VanillaNet~\cite{chen2023vanillanet}), and recent language model inspired models (AIM~\cite{el2024scalable}, VisionLLaMA~\cite{chu2024visionllama}). We present three observations: 1) The performance-parameter trade-off of iLLaMA surpasses other LM-inspired models such as AIM and VisionLLaMA, presumably due to its use of causal attention and soft mask training techniques. 2) iLLaMA exhibits a superior accuracy-throughput trade-off compared to strong hierarchical baselines such as ConvNeXt-V2-N/T/B and Swin-S/B. We attribute this to iLLaMA's isotropic design (each intermediate block has the same feature resolution), which benefits from a straightforward and efficient architecture, enhancing inference speed. 3) Scalability of model capacity and dataset size: After comprehensive pre-training on the expanded ImageNet-21K dataset, the iLLaMA-B model achieves more than $85.0\%$ accuracy on ImageNet-1K with under 100M parameters, significantly outperforming ViT-B's $84.0\%$. Upon scaling up to the larger iLLaMA-L, accuracy reaches $86.0\%$, exceeding that of ViT-L pre-trained on ImageNet-21K and the AIM-7B pre-trained on the DFN-2B+ dataset. To our knowledge, this showcases SOTA performance for LLaMA-type architectures. 

\begin{table*}[t]
\centering
\caption{ImageNet-1K accuracy. Throughput (images/s) are tested on Nvidia A100 GPU with a batch size of 1024. Hie.: Hierarchical, Iso.: Isotropic, Sup.: Supervised (pre-)training, AR.: Autoregressive pre-training. {\color{00blue}$\spadesuit$} ConvNet, {\color{02green}$\blacksquare$} Vision Transformer, 
{\color{orange}$\clubsuit$} MLP, {\color{02pink}$\maltese$} LM-inspired visual model, {\color{00red}$\bigstar$} LLaMA.}
\vspace{-0.7em}
\renewcommand{\arraystretch}{1}
\small
\label{tab:imagenet}
\setlength{\tabcolsep}{1.6pt}{
\begin{tabular}{l|c|c|c|c|c|c|c|c}
\toprule[1.5pt]
Model & Dataset Used & Objective & Type & Image Size & Params & MACs & Throughput & Acc \\
\midrule

{\color{00blue}$\spadesuit$} ConvNeXt-S~\cite{liu2022convnet}  & IN-1K & Sup. & Hie. & 224$\times$224 & 50M & 8.7G & 1185  & 83.1 \\
{\color{00blue}$\spadesuit$} ConvNeXt-B~\cite{liu2022convnet}  & IN-1K & Sup. & Hie. & 224$\times$224 & 89M & 15.4G & 877  & 83.8 \\
{\color{00blue}$\spadesuit$} ConvNeXt-L~\cite{liu2022convnet}  & IN-1K & Sup. & Hie. & 224$\times$224 & 198M & 34.4G & 543  & 84.3 \\
{\color{00blue}$\spadesuit$} ConvNeXtV2-N~\cite{woo2023convnext}  & IN-1K & Sup. & Hie. & 224$\times$224 & 15.6M & 2.45G & 2120  & 81.2 \\
{\color{00blue}$\spadesuit$} ConvNeXtV2-T~\cite{woo2023convnext}  & IN-1K & Sup. & Hie. & 224$\times$224 & 28.6M & 4.47G & 1362  & 82.5 \\
{\color{00blue}$\spadesuit$} ConvNeXtV2-B~\cite{woo2023convnext}  & IN-1K & Sup. & Hie. & 224$\times$224 & 88.7M & 15.4G & 645  & 84.3 \\

\midrule
{\color{02green}$\blacksquare$} Swin-S~\cite{liu2021swin}  & IN-1K & Sup. & Hie. & 224$\times$224 & 50M & 8.7G & 934 & 83.0 \\
{\color{02green}$\blacksquare$} Swin-B~\cite{liu2021swin}  & IN-1K & Sup. & Hie. & 224$\times$224 & 88M & 15.4G & 710 & 83.5 \\
{\color{02green}$\blacksquare$} DeiT-Ti~\cite{touvron2021training}  & IN-1K & Sup. & Iso. & 224$\times$224 & 5.7M & 1.3G & 6051 & 72.2 \\			
{\color{02green}$\blacksquare$} DeiT-S~\cite{touvron2021training}  & IN-1K & Sup. & Iso. & 224$\times$224 & 22.1M & 4.6G & 3080 & 79.8 \\
{\color{02green}$\blacksquare$} DeiT-B~\cite{touvron2021training}   & IN-1K & Sup. & Iso. & 224$\times$224 & 86.4M & 17.6G & 1348 & 81.8 \\
{\color{02green}$\blacksquare$} ViT-B/16~\cite{dosovitskiy2020image}  & IN-21K, IN-1K & Sup., Sup. & Iso. & 384$\times$384 & 86.4M & 55.5G & 349 & 84.0 \\
{\color{02green}$\blacksquare$} ViT-L/16~\cite{dosovitskiy2020image}  & IN-21K, IN-1K & Sup., Sup. & Iso. & 384$\times$384 & 304.1M & 191.2G & 124 & 85.2 \\

\midrule
{\color{orange}$\clubsuit$} PoolFormer-S12~\cite{yu2022metaformer}  & IN-1K & Sup. & Hie. & 224$\times$224 & 12M & 1.8G & 4354 & 77.2 \\
{\color{orange}$\clubsuit$} PoolFormer-M48~\cite{yu2022metaformer}  & IN-1K & Sup. & Hie. & 224$\times$224 & 73M & 11.6G & 768 & 82.5 \\
{\color{orange}$\clubsuit$} VanillaNet-5~\cite{chen2023vanillanet}  & IN-1K & Sup. & Hie. & 224$\times$224 & 15.5M & 5.2G & - & 72.5 \\
{\color{orange}$\clubsuit$} VanillaNet-13-1.5$\times$\cite{chen2023vanillanet}  & IN-1K & Sup. & Hie. & 224$\times$224 & 127.8M & 26.5G & - & 82.5 \\

\midrule
{\color{02pink}$\maltese$} AIM-0.6B~\cite{el2024scalable}  & DFN-2B+, IN-1K & AR., Sup. & Iso. & 224$\times$224 & 0.6B & - & - & 78.5 \\
{\color{02pink}$\maltese$} AIM-3B~\cite{el2024scalable}  & DFN-2B+, IN-1K & AR., Sup. & Iso. & 224$\times$224 & 3B & - & - & 82.2 \\
{\color{02pink}$\maltese$} AIM-7B~\cite{el2024scalable}  & DFN-2B+, IN-1K & AR., Sup. & Iso. & 224$\times$224 & 7B & - & - & 82.4 \\
{\color{02pink}$\maltese$} P-VisionLLaMA-S~\cite{chu2024visionllama}  & IN-1K & Sup. & Hie. & 224$\times$224 & 24M & - & - & 81.6 \\
{\color{02pink}$\maltese$} P-VisionLLaMA-B~\cite{chu2024visionllama}  & IN-1K & Sup. & Hie. & 224$\times$224 & 56M & - & - & 83.2 \\
{\color{02pink}$\maltese$} P-VisionLLaMA-L~\cite{chu2024visionllama}  & IN-1K & Sup. & Hie. & 224$\times$224 & 99M & - & - & 83.6 \\
{\color{02pink}$\maltese$} VisionLLaMA-L~\cite{chu2024visionllama}  & IN-1K, IN-1K & Sup., Sup. & Iso. & 224$\times$224 & 310M & - & - & 84.6 \\

\midrule	
{\color{00red}$\bigstar$} iLLaMA-T  & IN-1K & Sup. & Iso. & 224$\times$224 & 5.7M  & 1.3G & 6958 & 75.0  \\     
{\color{00red}$\bigstar$} iLLaMA-S  & IN-1K & Sup. & Iso. & 224$\times$224 & 21.9M  & 4.6G & 3222 & 79.9  \\
{\color{00red}$\bigstar$} iLLaMA-B  & IN-1K & Sup. & Iso. & 224$\times$224 & 86.3M  & 17.6G & 1345 & 81.6  \\
{\color{00red}$\bigstar$} iLLaMA-B  & IN-1K & Sup. & Iso. & 384$\times$384 & 86.3M  & 55.5G & 332 & 83.0  \\
{\color{00red}$\bigstar$} iLLaMA-B  & IN-21K, IN-1K & Sup., Sup. & Iso. & 224$\times$224 & 86.3M  & 17.6G & 1345 & 83.6  \\
{\color{00red}$\bigstar$} iLLaMA-B  & IN-21K, IN-1K & Sup., Sup. & Iso. & 384$\times$384 & 86.3M  & 55.5G & 332 & 85.0  \\
{\color{00red}$\bigstar$} iLLaMA-L  & IN-21K, IN-1K & Sup., Sup. & Iso. & 224$\times$224 & 310.2M  & 62.8G & 456 & 84.8  \\
{\color{00red}$\bigstar$} iLLaMA-L  & IN-21K, IN-1K & Sup., Sup. & Iso. & 384$\times$384 & 310.2M  & 194.7G & 116 & 86.0  \\
\bottomrule[1pt]
\end{tabular}
}
\vspace{-2.1em}
\end{table*}


\subsection{Model Calibration and Shape-Texture Bias} 
\label{sec:4.3}
\vspace{-0.5em}
Beyond ImageNet accuracy, we also examined iLLaMA's calibration properties and shape-texture bias for a more detailed evaluation, following~\cite{vishniakov2023convnet}. Besides iLLaMA, we also explore two prevalent architectures, \ie, ConvNeXt~\cite{liu2022convnet} and DeiT3~\cite{touvron2022deit}, representing ConvNets and Transformers, respectively. We apply ImageNet-21K pre-trained and ImageNet-1K fine-tuned models in this section. 

\textbf{Model calibration.} 
Model calibration represents the relationship between a model's precision and confidence across samples of varying difficulty, \ie, poor-calibrated models tend to produce overly confident yet incorrect predictions, whereas well-calibrated models demonstrate a strong correlation between confidence and accuracy~\cite{guo2017calibration}. Calibration is commonly measured using the Expected Calibration Error (ECE), where a lower ECE is favorable. ECE results for different models on ImageNet-1K are presented in Table~\ref{tab:beyond}. The calibration of iLLaMA is lower than that of DeiT3, suggesting a more reliable output confidence. We also plot the reliability diagrams~\cite{vishniakov2023convnet} to intuitively compare the calibration of different models, as shown in Figure~\ref{fig:calibration} of Appendix~\ref{sec:8.7}.

\textbf{Shape-texture bias.} Shape-texture bias measures the extent to which the model relies on the shape or texture of the image when performing recognition~\cite{geirhos2018imagenet}. We generally prefer models to mimic human eye behavior, relying more on shape rather than texture~\cite{tuli2021convolutional, geirhos2020shortcut}. We calculate the shape ratio for all models on cue-conflict images and report the results in Table~\ref{tab:beyond}, following~\cite{vishniakov2023convnet}. Our iLLaMA shows the largest shape ratio of $41.45\%$ among the three compared baselines, suggesting the potential of the LLM architecture for vision. Detailed results are provided in Figure~\ref{fig:shape_texture_bias} of Appendix~\ref{sec:8.8}.

\subsection{Compatibility with Quantization} 
\label{sec:4.4}

Since a practical goal for neural networks is deployment on low-bit hardware chips, we further examine iLLaMA's compatibility with quantization. We basically follow Q-ViT~\cite{li2022q} to apply quantization-aware training (QAT) to iLLaMA, with weights and activations of all blocks' FFN and causal self-attention layers to 8 bits. Quantization recipes and results are shown in Table~\ref{tab:setup_quantization} of Appendix~\ref{subsec:8.3.4} and Table~\ref{tab:quant}. Different sizes of low-bit iLLaMA maintain accuracy well, and 8-bit iLLaMA-T is even compete favorably with the full-precision DeiT-T~\cite{touvron2021training} ($72.4\%$ \text{v.s.} $72.2\%$).

\begin{table}[t]
        \vspace{-1.2em}
	\begin{minipage}[t]{0.4\linewidth}
		\centering
        \small
		\renewcommand{\arraystretch}{1.06} 
		\caption{Quantization results. $\#$Bits ($\mathrm{w}$-$\mathrm{a}$): $\mathrm{w}$ bit weights, $\mathrm{a}$ bit activations. 8-bit iLLaMA-T matches 32-bit DeiT-T.}\label{tab:quant}
		\vspace{0.15em}
		\setlength{\tabcolsep}{8pt}
            \begin{tabular}{lcccc}
                \toprule[1.5pt]
                Model & $\#$Bits & Tiny & Small \\
                \midrule
                DeiT~\cite{touvron2021training} & 32-32 & 72.2 & 79.8 \\
                iLLaMA & 32-32 & 75.0 & 79.9 \\
                \gr iLLaMA & 8-8 & 72.4 & 77.4 \\
                \bottomrule[1pt]
            \end{tabular}
	\end{minipage}
	\hspace{.15in}
	\begin{minipage}[t]{0.56\linewidth}
		\centering
        \small
		\renewcommand{\arraystretch}{1.06} 
		\caption{Calibration (expected calibration error~$\downarrow$) and shape-texture bias (ratio~$\uparrow$) results of ConvNeXt-B~\cite{liu2022convnet}, DeiT3-B~\cite{touvron2022deit} and iLLaMA-B. We use both IN-21K pre-trained and IN-1K fine-tuned models.}\label{tab:beyond}
		\vspace{0.1em}
		\setlength{\tabcolsep}{3.7pt}
            \begin{tabular}{lccc}
            \toprule[1.5pt]
            Evaluation & ConvNeXt-B & DeiT3-B & iLLaMA-B \\
            \midrule
            Calibration & 0.0281 & 0.0415 & 0.0335 \\
            Shape-Texture Bias & 33.30$\%$ & 39.86$\%$ & 41.45$\%$ \\
            \bottomrule[1pt]
            \end{tabular}
	\end{minipage}
	\vspace{-0.em}
\end{table}


\subsection{Transferability on Downstream Tasks} 
\label{sec:4.5}

\textbf{CIFAR transfer learning.} We fine-tune ViT-T and iLLaMA-T on the CIFAR datasets~\cite{krizhevsky2009learning}, including an ablation of the soft mask on iLLaMA. Detailed recipes are shown in Table~\ref{tab:setup_transfer} of Appendix~\ref{subsec:8.3.5}. iLLaMA's performance on CIFAR datasets is essentially on par with ViT, assuring that iLLaMA can be confidently applied in the transfer learning field as a practical alternative to ViT. Additionally, soft mask is helpful in the relatively complicated CIFAR100, demonstrating its generalizability. 

\textbf{ADE20K semantic segmentation.} We fine-tune our ImageNet-1K pre-trained iLLaMA and ViT models on ADE20K~\cite{zhou2019semantic} dataset using UperNet~\cite{xiao2018unified} to perform semantic segmentation task. For both iLLaMA and ViT, we set the learning rate as 6e-5 and weight decay as 0.01. Table~\ref{tab:seg} presents the results. iLLaMA's performance is marginally lower than ViT's, which we attribute to the potential impact of the masking mechanism in iLLaMA's causal attention on high-resolution dense prediction tasks. This suggests there is still space for optimization, a subject for future investigation.   

\begin{table}[t]
        \vspace{-1.3em}
	\begin{minipage}[t]{0.37\linewidth}
		\centering
        \small
		\renewcommand{\arraystretch}{1.1} 
		\caption{Soft mask for CIFAR transfer learning. Soft mask improves iLLaMA performance without changing the inference architecture.}\label{tab:cifar}
		\vspace{0.1em}
		\setlength{\tabcolsep}{5.3pt}
            \begin{tabular}{lcc}
                \toprule[1.5pt]
                Model & CIFAR10 & CIFAR100 \\
                \midrule
                ViT-T & 98.0 & 85.5 \\
                iLLaMA-T & 97.9 & 84.8 \\
                \gr + soft mask &  97.9 & 85.5 \\
                \bottomrule[1pt]
            \end{tabular}
	\end{minipage}
	\hspace{.15in}
	\begin{minipage}[t]{0.59\linewidth}
		\centering
        \small
		\renewcommand{\arraystretch}{1.1} 
		\caption{ADE20K semantic segmentation results using UperNet~\cite{xiao2018unified}. We report mIoU with multi-scale testing.  FLOPs calculation are based on input sizes of (512, 512).}\label{tab:seg}
		\vspace{0.1em}
		\setlength{\tabcolsep}{7pt}
            \begin{tabular}{lcccc}
            \toprule[1.5pt]
            Backbone & Input Crop. & mIoU & \#Param. & FLOPs  \\
            \midrule
            ViT-T & 512$^2$ & 39.8 & 10.88M & 37.1G  \\
            \gr
            iLLaMA-T &  512$^2$ & 37.7 & 10.86M & 37.1G \\
            ViT-B & 512$^2$ & 47.3 & 163.29M & 585.7G  \\
            \gr
            iLLaMA-B &  512$^2$ & 45.1 & 163.22M &  585.7G \\
            \bottomrule[1pt]
            \end{tabular}
	\end{minipage}
	\vspace{-1.5em}
\end{table}


\section{Conclusions}
\label{sec:conclusions}
In this paper, we systematically studies whether Transformer decoder, an architecture that has shown amazing potential in LLMs, can also take root in learning visual representation through straightforward supervised training. The key component -- causal self-attention we used -- is not novel and is inherited from existing LLM architectures, but we propose pivotal techniques, \ie, PS [cls] and soft mask strategies, to effectively adapt them to visual tasks. The proposed iLLaMA outperforms many ConvNets, ViTs, and MLPs on imagenet, and demonstrates robust quantization compatibility, calibration, and shape-texture bias, thereby showing its practicality. We hope that this work will inspire a rethinking of generic yet practical architecture that can fully unify both vision and text.

\medskip

{\small 
	\bibliographystyle{plain}
	\bibliography{main}
}


\appendix


\section{Network Configuration}
\label{sec:8.1}
In Table~\ref{tab:hyper}, we provide detailed architecture configurations for iLLaMA models of various capacities. Our approach to scaling up the model size, from small to large, is similar to that of the ViT. Thus, akin to ViT, iLLaMA benefits from the simplicity of an isotropic architecture and high throughput, with its internal features remaining unchanged in resolution and number of channel as the depth increases. 

\begin{wraptable}{r}{0.63\textwidth}
\vspace{-1.2em}
\centering
\small
\caption{Detailed iLLaMA architecture configurations.}
\label{tab:hyper}
\begin{tabular}{l|c|c|c|c}
\toprule[1.5pt]
& Tiny (T) & Small (S) & Base (B) & Large (L) \\
\midrule
    depth & 12 & 12 & 12 & 24 \\
    embedding dim & 192 & 384 & 768 & 1024 \\
    number of heads & 3 & 6 & 12 & 16 \\
    \#param. (M) & 5.7 & 21.9 & 86.3 & 310.2 \\
    MACs (G) & 1.3 & 4.6 & 17.6 & 62.8 \\
\bottomrule
\end{tabular}
\end{wraptable}

\begin{figure*}[h]
	\centering
	\includegraphics[width=0.75\linewidth]{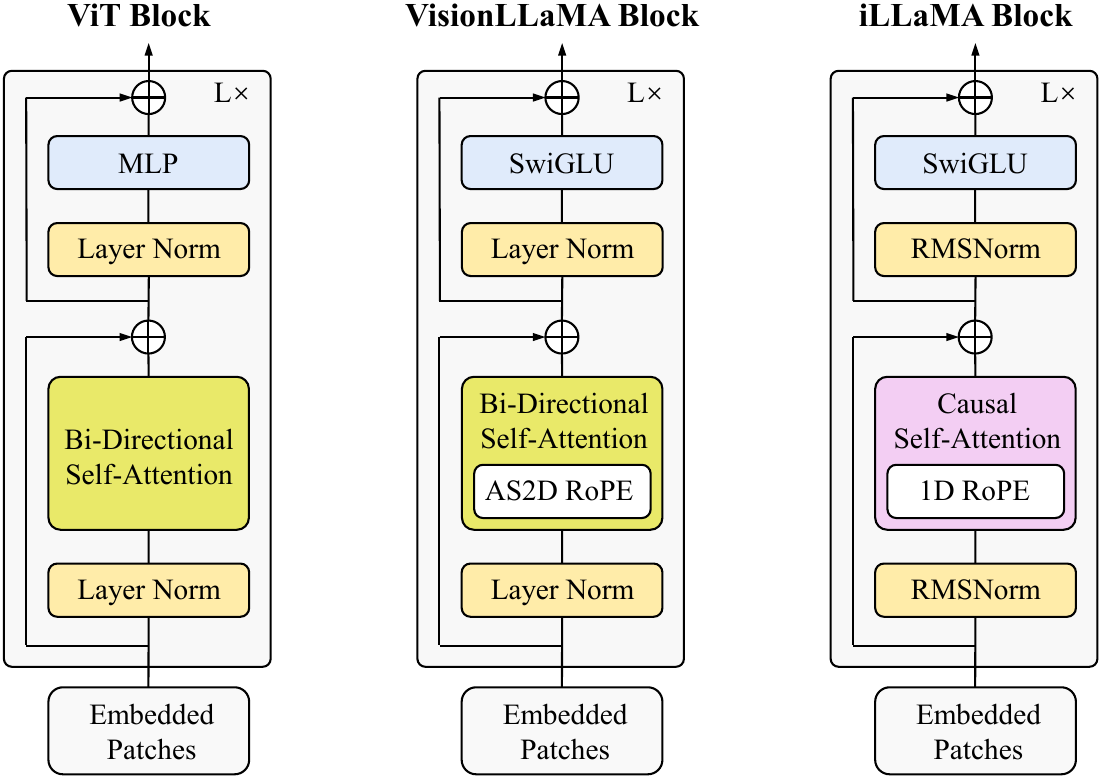}
	\caption{Comparison between ViT~\cite{dosovitskiy2020image}, VisionLLaMA~\cite{chu2024visionllama}, and iLLaMA blocks. }
	\label{fig:comparison}
\end{figure*}

We provide a block-level comparison between iLLaMA and ViT model in Figure~\ref{fig:comparison}. 
VisionLLaMA uses SwiGLU, and AS2D RoPE to build LLaMA-style architecture. Differently, we further uses RMSNorm, modified causal self-attention and 1D RoPE from LLaMA to replace layer normalization, bi-directional self-attention, and proposes two pivotal strategies, \ie, \textit{PS [cls]} and \textit{soft mask} to help the optimization of our iLLaMA. We also keep the learnable positional embedding as ViT does.

\section{PyTorch-like Code of iLLaMA Causal Self-attention}
\label{sec:8.2}

The PyTorch-like implementation of our iLLaMA causal self-attention is shown as Algorithm~\ref{alg:illama_block}. 
The iLLaMA code exhibits a high degree of similarity in structure and composition to the official LLaMA code~\footnote{\url{https://github.com/meta-llama/llama}} released by Meta, potentially offering considerable coding cost savings in developing a unified vision and language network with such architecture.

\begin{algorithm}[h]
\caption{PyTorch code of iLLaMA causal self-attention}
\label{alg:illama_block}
\definecolor{codeblue}{rgb}{0.25,0.5,0.5}
\definecolor{codekw}{rgb}{0.85, 0.18, 0.50}
\lstset{
  backgroundcolor=\color{white},
  basicstyle=\fontsize{7.5pt}{7.5pt}\ttfamily\selectfont,
  columns=fullflexible,
  breaklines=true,
  captionpos=b,
  commentstyle=\fontsize{7.5pt}{7.5pt}\color{codeblue},
  keywordstyle=\fontsize{7.5pt}{7.5pt}\color{codekw},
}
\begin{lstlisting}[language=python]
import torch
import torch.nn as nn

def reshape_for_broadcast(freqs_cis: torch.Tensor, x: torch.Tensor):
    ndim = x.ndim  
    assert 0 <= 1 < ndim
    assert freqs_cis.shape == (x.shape[1], x.shape[-1])  
    shape = [d if i == 1 or i == ndim - 1 else 1 for i, d in enumerate(x.shape)]  
    return freqs_cis.view(*shape)  

def apply_rotary_emb(
    xq: torch.Tensor,
    xk: torch.Tensor,
    freqs_cis: torch.Tensor,
) -> Tuple[torch.Tensor, torch.Tensor]:
    xq_ = torch.view_as_complex(xq.float().reshape(*xq.shape[:-1], -1, 2))  
    xk_ = torch.view_as_complex(xk.float().reshape(*xk.shape[:-1], -1, 2))  
    freqs_cis = reshape_for_broadcast(freqs_cis, xq_)  
    xq_out = torch.view_as_real(xq_ * freqs_cis).flatten(3)  
    xk_out = torch.view_as_real(xk_ * freqs_cis).flatten(3)  
    return xq_out.type_as(xq), xk_out.type_as(xk)  


class Attention(nn.Module):
    def __init__(self, dim, num_heads=8, qkv_bias=False, qk_scale=None, attn_drop=0., proj_drop=0.):
        super().__init__()
        self.num_heads = num_heads
        head_dim = dim // num_heads
        # NOTE scale factor was wrong in my original version, can set manually to be compat with prev weights
        self.scale = qk_scale or head_dim ** -0.5

        self.qkv = nn.Linear(dim, dim * 3, bias=qkv_bias)
        self.proj = nn.Linear(dim, dim)

    def forward(self, x: torch.Tensor, freqs_cis: torch.Tensor, mask: Optional[torch.Tensor]):
        B, N, C = x.shape
        qkv = self.qkv(x).reshape(B, N, 3, self.num_heads, C // self.num_heads).permute(2, 0, 1, 3, 4) # [3, B, N, self.num_heads, C // self.num_heads]
        q, k, v = qkv[0], qkv[1], qkv[2]   # make torchscript happy (cannot use tensor as tuple) # [B, N, self.num_heads, C // self.num_heads]

        q, k = apply_rotary_emb(q, k, freqs_cis=freqs_cis)

        q = q.transpose(1, 2)  # [B, self.num_heads, N, C // self.num_heads]
        k = k.transpose(1, 2)  # [B, self.num_heads, N, C // self.num_heads]
        v = v.transpose(1, 2)  # [B, self.num_heads, N, C // self.num_heads]
        attn = (q @ k.transpose(-2, -1)) * self.scale # [B, self.num_heads, N, N]
        attn = attn.softmax(dim=-1)
        if mask is not None:
            attn = attn * mask  # (B, H, N, N)

        x = (attn @ v).transpose(1, 2).reshape(B, N, C)
        x = self.proj(x)
        
        return x
        
\end{lstlisting}
\end{algorithm}

\section{Experimental Settings}
\label{sec:8.3}
\subsection{Training Recipe in Section~\ref{sec:method_vision}} 
\label{subsec:8.3.1}
Our training recipe for training the tiny and base models during the ``designing iLLaMA: a roadmap'' (Section~\ref{sec:method_vision}) is primarily adapted from ConvNeXt~\cite{liu2022convnet, liu2023dropout}, summarized in Table~\ref{tab:setup_basic}. 

Basically, both regimes use the same experimental setup, with the only difference being the stochastic depth rate at 0.0 and 0.4, respectively. Notably, for the ViT baseline, our experimental results are $73.8\%$ and $81.5\%$, as shown in Table~\ref{tab:exploration}, which slightly differ from the results of $73.9\%$ and $81.6\%$ reported in~\cite{liu2023dropout}. 

Utilizing only the basic training recipe with architectural modifications, the performance of iLLaMA's tiny and base models achieves $73.2\%$ and $81.2\%$, as shown in Table~\ref{tab:exploration}, yet remains below the ViT baseline. We attribute this to the impairing effect of causal self-attention on the information mixing among tokens. Thus, we enhance the training recipe, detailed next. 

\begin{table*}[h]
\centering
\small
\caption{Our training recipe for Section 3 in the main paper, adapted from~\cite{liu2023dropout}.}
\label{tab:setup_basic}
\begin{tabular}{l|c} 
\toprule[1.5pt]
Training Configuration & iLLaMA-T/B \\
\midrule
\textit{Initialization:} & \\
weight init & trunc. normal (0.2)  \\
\midrule
\textit{Training recipe:} & \\
optimizer & AdamW \cite{loshchilov2017decoupled} \\
optimizer momentum & $\beta_1, \beta_2{=}0.9, 0.999$  \\
\midrule
\textit{Learning hyper-parameters:} & \\
base learning rate & 4e-3  \\
learning rate schedule & cosine decay  \\
weight decay & 0.05  \\
batch size & 4096  \\
training epochs & 300  \\
lr warmup epochs & 50  \\
warmup schedule & linear  \\
gradient clip & None  \\
exp. mov. avg. (EMA) \cite{polyak1992acceleration} & None \\
\midrule
\textit{Dropout:} & \\
dropout rate \cite{hinton2012improving} & 0.0 \\
stochastic depth rate \cite{huang2016deep} & 0.0/0.4 \\
\midrule
\textit{Data augmentation:} & \\
input resolution & $224^2$ \\
randAugment \cite{cubuk2020randaugment} & (9, 0.5)  \\
random erasing \cite{zhong2020random} & 0.25  \\
label smoothing \cite{szegedy2016rethinking} & 0.1  \\
mixup \cite{zhang2017mixup} & 0.8  \\
cutmix \cite{yun2019cutmix} & 1.0  \\
\bottomrule
\end{tabular}
\end{table*}

\subsection{ImageNet (Pre-)training Recipe} 
\label{subsec:8.3.2}
As illustrated in Table~\ref{tab:setup_pretrain}, we provide the detailed ImageNet-1K training hyper-parameters and ImageNet-21K pre-training hyper-parameters for the experimental results in Table~\ref{tab:imagenet}. 

For the iLLaMA-T/S/B models, we train directly on ImageNet-1K and discover that models of different sizes are suited to different soft mask settings. For instance, the soft mask schedules are set to constant/linear/linear, respectively, with cutoff epochs designated as 50/50/25. We train the iLLaMA-T/S/B models using 8 A100 GPUs.

We pre-trained the iLLaMA-B/L models on ImageNet-21K for 90 epochs, adhering to the practices in~\cite{liu2022convnet}. We set the cutoff epochs to 30, indicating that the iLLaMA models' self-attention fully transitions to causal self-attention after 30 epochs. We pre-train the iLLaMA-B/L models using 8 A100 GPUs.

\begin{table*}[h]
\centering
\small
\caption{Our (pre-)training settings for iLLaMa model on ImageNet-1K/ImageNet-21K, respectively, adapted from~\cite{liu2023dropout}. Some key training techniques are \colorbox{DecGrey}{highlighted}.}
\label{tab:setup_pretrain}
\begin{tabular}{@{\hskip 1.5ex}l|c@{\hskip 3ex}c}
\toprule[1.5pt]
& iLLaMA-T/S/B & iLLaMA-B/L \\
\multirow{1}{*}{(Pre-)Training Configuration} & ImageNet-1K & ImageNet-21K \\
\midrule
\textit{Initialization:} & \\
weight init & trunc. normal (0.2)  & trunc. normal (0.2) \\
\midrule
\textit{Training recipe:} & \\
optimizer & AdamW & AdamW \\
optimizer momentum & $\beta_1, \beta_2{=}0.9, 0.999$ & $\beta_1, \beta_2{=}0.9, 0.999$  \\
\midrule
\textit{Learning hyper-parameters:} & \\
base learning rate & 4e-3 & 1e-3 \\
learning rate schedule & cosine decay & cosine decay \\
weight decay & 0.05 & 0.01 \\
batch size & 4096 & 4096 \\
training epochs & 300 & 90 \\
warmup schedule & linear & linear \\
gradient clip & None & None \\
exp. mov. avg. (EMA) & None & None \\
\midrule
\textit{Dropout:} & \\
dropout rate & 0.0 & 0.0 \\
\gr stochastic depth rate & 0.0/0.1/0.4 & 0.1 \\
\midrule
\textit{Data augmentation:} & \\
input resolution & $224^2$ & $224^2$ \\
randAugment & (9, 0.5) & (9, 0.5) \\
random erasing & 0.25 & 0.25 \\
label smoothing & 0.1 & 0.1 \\
\gr mixup  & 0.1/0.5/0.95 & 0.8 \\
\gr cutmix & 0.1/0.5/1.0 & 1.0 \\
\midrule
\textit{Soft mask:} & \\
\gr soft mask schedule & constant/linear/linear & constant \\
\gr cutoff epochs & 50/50/25 & 30 \\
\gr lr warmup epochs & 5/5/50 & 5 \\
\bottomrule
\end{tabular}
\end{table*}


\subsection{ImageNet Fine-tuning Recipe} 
\label{subsec:8.3.3}
We present the results of fine-tuning models pre-trained on ImageNet-1K at a resolution of $384\times384$, as well as the outcomes of fine-tuning models pre-trained on ImageNet-21K at resolutions of $224\times224$ and $384\times384$, as shown in Table~\ref{tab:setup_finetune}. All ImageNet-1K fine-tuning experiments were conducted for 30 epochs, following the convention in~\cite{liu2022convnet}. 

For the iLLaMA-B model pre-trained on ImageNet-1K, we used a relatively higher stochastic depth rate of 0.8. For the iLLaMA-B/L models pre-trained on ImageNet-21K, we employed relatively lower stochastic depth rates of 0.2 and 0.3, respectively. 

Additionally, we standardized the cutoff epoch at 0 for our ImageNet-1K fine-tuning experiments, ensuring the application of a causal mask in self-attention to align with the LLaMA architecture. We also opted not to use learning rate warmup. We fine-tune the models using 8 A100 GPUs. 

\begin{table*}[h]
\centering
\small
\caption{Our fine-tuning settings for iLLaMa model on ImageNet-1K, adapted from~\cite{liu2023dropout}. Some key training techniques are \colorbox{DecGrey}{highlighted}.}
\label{tab:setup_finetune}
\begin{tabular}{@{\hskip 1.5ex}l|c@{\hskip 3ex}c@{\hskip 3ex}c}
\toprule[1.5pt]
& iLLaMA-B & iLLaMA-B/L & iLLaMA-B/L \\
\multirow{2}{*}{(Pre-)Training Configuration} & ImageNet-1K   & ImageNet-21K  & ImageNet-21K  \\
 &  224$^2$  & 224$^2$  & 224$^2$  \\
\midrule
\multirow{1}{*}{Fine-Tuning Configuration} & ImageNet-1K & ImageNet-1K & ImageNet-1K \\
\midrule
\textit{Initialization:} & \\
weight init & trunc. normal (0.2)  & trunc. normal (0.2)  & trunc. normal (0.2) \\
\midrule
\textit{Training recipe:} & \\
optimizer & AdamW & AdamW & AdamW \\
optimizer momentum & $\beta_1, \beta_2{=}0.9, 0.999$ & $\beta_1, \beta_2{=}0.9, 0.999$  & $\beta_1, \beta_2{=}0.9, 0.999$  \\
\midrule
\textit{Learning hyper-parameters:} & \\
base learning rate & 8e-5 & 8e-5/6e-5 & 1.1e-4/3.5e-5 \\
learning rate schedule & cosine decay & cosine decay & cosine decay \\
weight decay & 1e-8 & 1e-8 & 1e-8 \\
batch size & 512 & 512 & 512 \\
training epochs & 30 & 30 & 30 \\
warmup schedule & linear & linear & linear \\
gradient clip & None & None & None \\
exp. mov. avg. (EMA) & None & None & None \\
\midrule
\textit{Dropout:} & \\
dropout rate & 0.0 & 0.0 & 0.0 \\
\gr stochastic depth rate & 0.8 & 0.2/0.3 & 0.2/0.3 \\
\midrule
\textit{Data augmentation:} & \\
input resolution & $384^2$ & $224^2$ & $384^2$ \\
randAugment & (9, 0.5) & (9, 0.5) & (9, 0.5) \\
random erasing & 0.25 & 0.25 & 0.25 \\
label smoothing & 0.1 & 0.1 & 0.1 \\
\gr mixup  & 0 & 0 & 0 \\
\gr cutmix & 0 & 0 & 0 \\
\midrule
\textit{Soft mask:} & \\
\gr soft mask schedule & constant & constant & constant \\
\gr cutoff epochs & 0 & 0 & 0 \\
\gr lr warmup epochs & 0 & 0 & 0 \\
\bottomrule
\end{tabular}
\end{table*}


\subsection{Quantization-aware Training Recipe}
\label{subsec:8.3.4}
We provide our quantization-aware training recipe for iLLaMA in Table~\ref{tab:setup_quantization}. Basically we follow the Q-ViT method proposed in~\cite{li2022q}, with only weights and activations in each basic block's causal self-attention and FFN module are quantized to 8 bit width. 

\begin{table*}[t]
\centering
\small
\caption{Our quantization-aware training settings for iLLaMa model on ImageNet-1K, adapted from~\cite{liu2023dropout, li2022q}. Some key training techniques are \colorbox{DecGrey}{highlighted}.}
\label{tab:setup_quantization}
\begin{tabular}{@{\hskip 1.5ex}l|c}
\toprule[1.5pt]
& iLLaMA-T/S \\
\multirow{1}{*}{(Pre-)Training Configuration} & ImageNet-1K \\
\midrule
\textit{Initialization:} & \\
weight init & trunc. normal (0.2) \\
\midrule
\textit{Training recipe:} & \\
optimizer & AdamW \\
optimizer momentum & $\beta_1, \beta_2{=}0.9, 0.999$ \\
\midrule
\textit{Learning hyper-parameters:} & \\
base learning rate & 3e-3/4e-3 \\
learning rate schedule & cosine decay \\
weight decay & 0.05 \\
batch size & 4096 \\
training epochs & 300 \\
warmup schedule & linear \\
gradient clip & None \\
exp. mov. avg. (EMA) & None \\
\midrule
\textit{Dropout:} & \\
dropout rate & 0.0 \\
\gr stochastic depth rate & 0.0/0.1 \\
\midrule
\textit{Data augmentation:} & \\
input resolution & $224^2$ \\
randAugment & (9, 0.5) \\
random erasing & 0.25 \\
label smoothing & 0.1 \\
\gr mixup  & 0.1/0.5 \\
\gr cutmix & 0.1/0.5 \\
\midrule
\textit{Soft mask:} & \\
\gr soft mask schedule & constant/linear \\
\gr cutoff epochs & 50/50 \\
\gr lr warmup epochs & 5/5 \\
\bottomrule
\end{tabular}
\end{table*}


\subsection{CIFAR Transfer Learning Recipe}
\label{subsec:8.3.5} 
We further provide our training recipe for transfer learning on the CIFAR10 and CIFAR100 datasets, as shown in Table~\ref{tab:setup_transfer}. 

In our transfer learning experiments, we consistently apply a linear soft mask schedule. However, for the CIFAR10 and CIFAR100 datasets, we use cutoff epochs of 25 and 50, respectively. 

\begin{table*}[h]
\centering
\small
\caption{Our transfer learning settings for ViT-T and iLLaMa-T  model on CIFAR10/100, respectively, adapted from~\cite{xu2023initializing}. Some key training techniques are \colorbox{DecGrey}{highlighted}.}
\label{tab:setup_transfer}
\begin{tabular}{@{\hskip 1.5ex}l|c@{\hskip 3ex}c}
\toprule[1.5pt]
\multirow{1}{*}{Transfer Learning Configuration} & CIFAR10 & CIFAR100 \\
\midrule
\textit{For both ViT-T and iLLaMA-T:} & \\
base learning rate & 2e-3 & 2e-3 \\
batch size & 1024 & 1024  \\
training epochs & 300 & 300  \\
stochastic depth rate & 0.0 & 0.0  \\
lr warmup epochs & 50 & 50  \\
\midrule
\textit{For iLLaMA-T only:} & \\
\gr soft mask schedule  & linear & linear \\
\gr cutoff epochs  & 25 & 50 \\ 
\bottomrule 
\end{tabular} 
\end{table*} 


\begin{figure*}[t]
	\centering
	\includegraphics[width=0.68\linewidth]{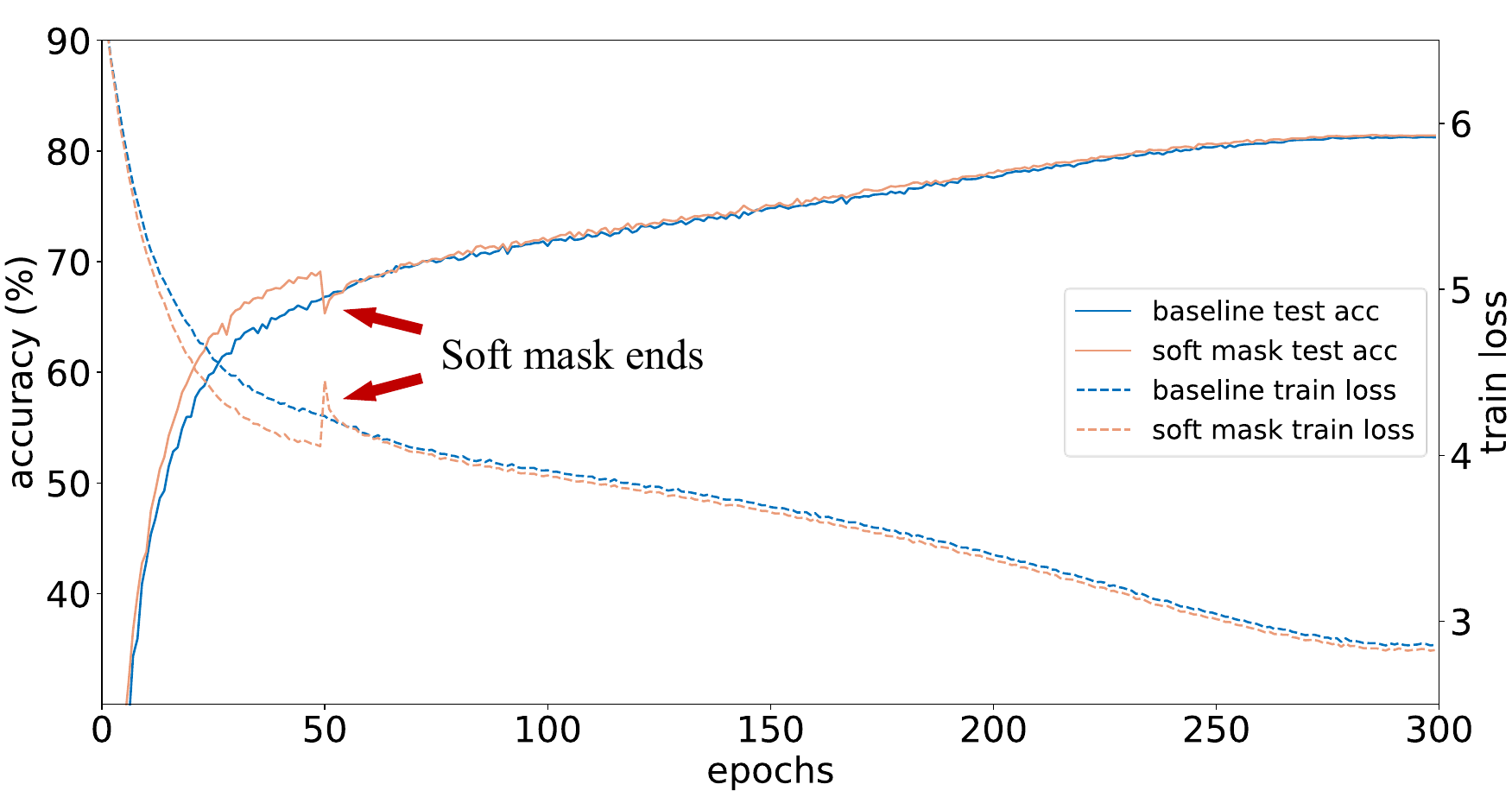}
	\caption{Training curves for iLLaMA-B/16 regime w/ and w/o soft mask. When soft mask ends, the model experiences a similar pattern to the training curve of iLLaMA-T/16 regime, with eventually a lower test loss observed.}
	\label{fig:train_loss_base}
\end{figure*}

\section{Designing iLLaMA: detailed results}
\label{sec:8.4}
We present the comprehensive experimental results of our exploration journey of iLLaMA in Table~\ref{tab:exploration}. This table not only delineates the stepwise accuracy of both the tiny and base models, as depicted in Figure~\ref{fig:architecture}, but also outlines the training loss at each step. The general trend observed is that as the training loss of the models decreases, their accuracy increases. 

Overall, the trend in changes for the base model is broadly similar to that of the tiny model. However, in contrast to the tiny model, the implementation of RoPE coupled with subsequent integration of LPE does not affect the base model's performance. This lack of impact, we theorize, stems from the base regime's reduced susceptibility to underfitting compared to the tiny regime, hence the addition of extra learnable parameters offers less benefit to its performance. 

Notably, vanilla causal self-attention mechanism proves inadequate for model optimization, an issue effectively addressed by implementing the PS [CLS] method. Additionally, the application of the soft mask technique significantly contributes to the training efficacy of both model sizes. 
\begin{table*}[h]
\centering
\small
\addtolength{\tabcolsep}{-2.0pt}
\def\arraystretch{1.18}
\caption{
ImageNet-1K classification accuracy via gradually replacing components in ViT-T/16 and ViT-B/16 with counterparts in LLaMA, \hlg{better} or \hlr{worse} than the ViT baseline results with our basic training recipe. Components from or modified from LLaMA are \colorbox{DecGrey}{highlighted}. 
P.E.: positional embedding, Bd.: bi-directional self-attention, Cs.: causal self-attention.
}
\label{tab:exploration}
\begin{tabular}{l|llll|cccccc}
\toprule[1.5pt]
Ablation & FFN & Norm & Attention & P.E. & Tiny & Train Loss & Base & Train Loss \\
\midrule
ViT~\cite{touvron2021training}  & MLP & LN & Bd. & LPE & 72.2 & - & 81.8 & - \\
\multicolumn{7}{l}{\scriptsize{results with our basic training recipe}}   \\
ViT & MLP & LN & Bd. & LPE  & 73.8 & 3.451 & 81.5 & 2.828  \\
+ LLaMa FFN    & \llamagrey{SwiGLU} & LN & Bd. & LPE  & \hlg{74.3}  & 3.407 & \hlg{82.0} & 2.724 \\
+ LLaMa Norm    & \llamagrey{SwiGLU} & \llamagrey{RMS} & Bd. & LPE  & \hlg{74.5} & 3.406 & \hlg{81.7} & 2.721  \\
+ LLaMa S.A. & \llamagrey{SwiGLU} & \llamagrey{RMS} & \llamagrey{Cs.} & LPE  & \hlr{0.1}  & Failed & \hlr{0.1} & Failed \\
+ LLaMa S.A.  & \llamagrey{SwiGLU} & \llamagrey{RMS} & \llamagrey{Cs. + \textit{\textbf{PS [CLS]}}} & LPE  & \hlr{71.9} & 3.599 & \hlr{80.6} & 2.869 \\
+ LLaMa P.E. & \llamagrey{SwiGLU} & \llamagrey{RMS} & \llamagrey{Cs. + \textit{\textbf{PS [CLS]}}} & \llamagrey{RoPE}  & \hlr{72.6} & 3.618 & \hlr{81.2} & 2.861  \\
+ LPE P.E. & \llamagrey{SwiGLU} & \llamagrey{RMS} & \llamagrey{Cs. + \textit{\textbf{PS [CLS]}}} & \llamagrey{RoPE} + LPE & \hlr{73.2} & 3.531 & \hlr{81.2} & 2.839  \\

\multicolumn{7}{l}{\scriptsize{modify the training techniques}}   \\
+ data aug. & \llamagrey{SwiGLU} & \llamagrey{RMS} & \llamagrey{Cs. + \textit{\textbf{PS [CLS]}}} & \llamagrey{RoPE} + LPE  & \hlg{74.3} & 2.990  & \hlr{81.3} & 2.868 \\
+ \textbf{\textit{soft mask}} & \llamagrey{SwiGLU} & \llamagrey{RMS} & \llamagrey{Cs. + \textit{\textbf{PS [CLS]}}} & \llamagrey{RoPE} + LPE &  \hlg{\textbf{75.0}}  & 2.955  & \hlg{\textbf{81.6}} & 2.828 \\

\bottomrule[1pt]
\end{tabular}
\scriptsize
\end{table*}

\section{Rank Analysis of causal Self-attention}
\label{sec:8.5}
\begin{figure*}[t]
    \centering

    \begin{subfigure}{0.3\textwidth}
        \includegraphics[width=\linewidth]{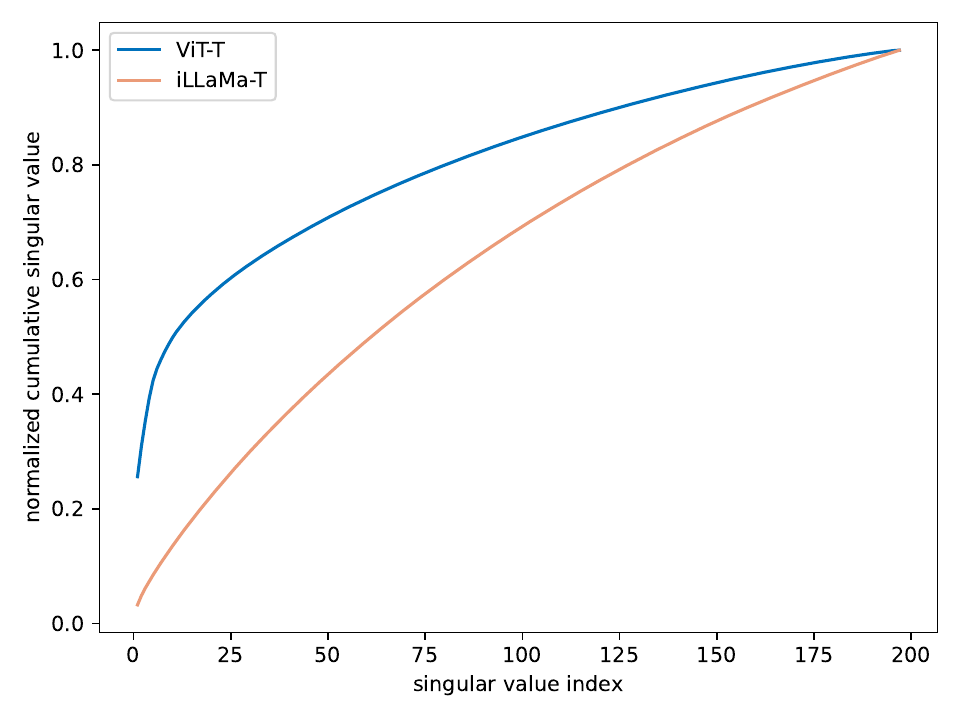}
        \caption{layer 1, head 1}
    \end{subfigure}
    \hspace{0.02\textwidth}
    \begin{subfigure}{0.3\textwidth}
        \includegraphics[width=\linewidth]{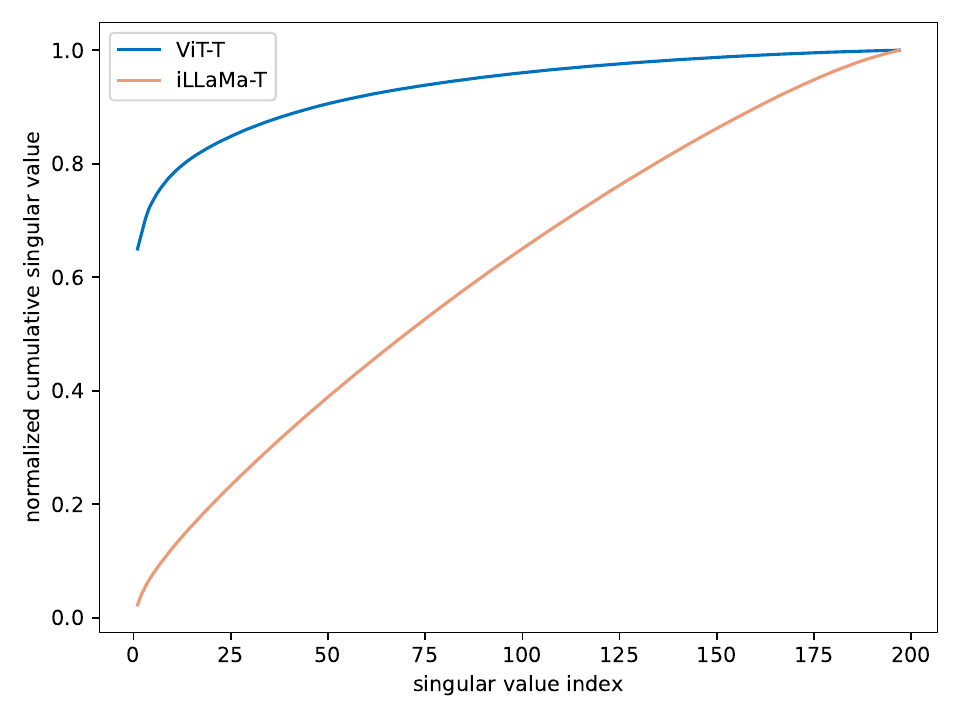}
        \caption{layer 1, head 2}
    \end{subfigure}
    \hspace{0.02\textwidth}
    \begin{subfigure}{0.3\textwidth}
        \includegraphics[width=\linewidth]{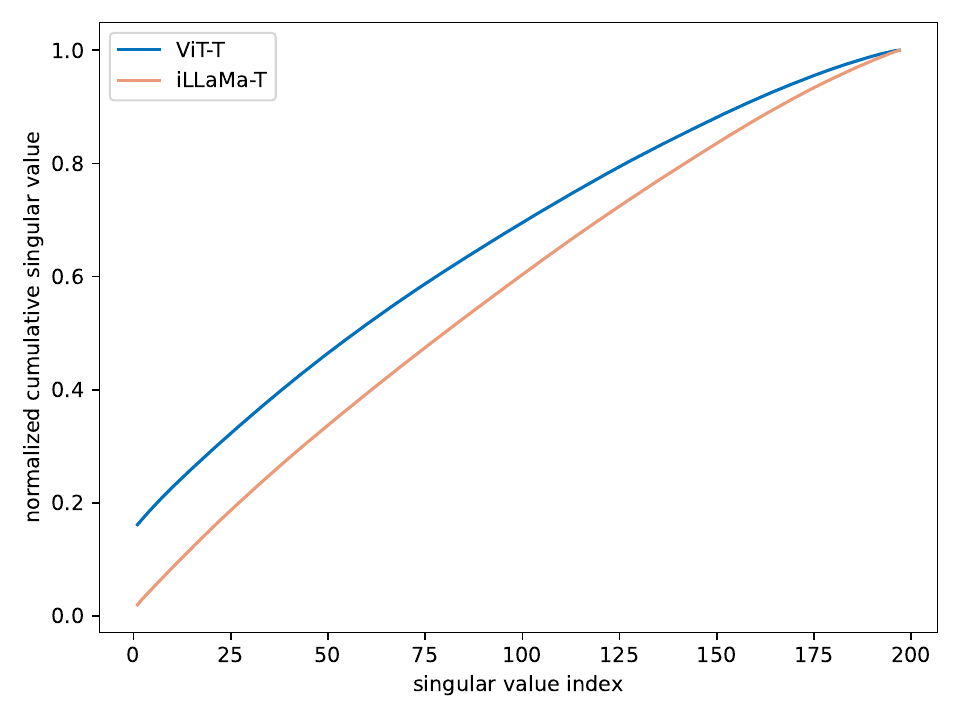}
        \caption{layer 1, head 3}
    \end{subfigure}

    \vspace{0.5cm}

    \begin{subfigure}{0.3\textwidth}
        \includegraphics[width=\linewidth]{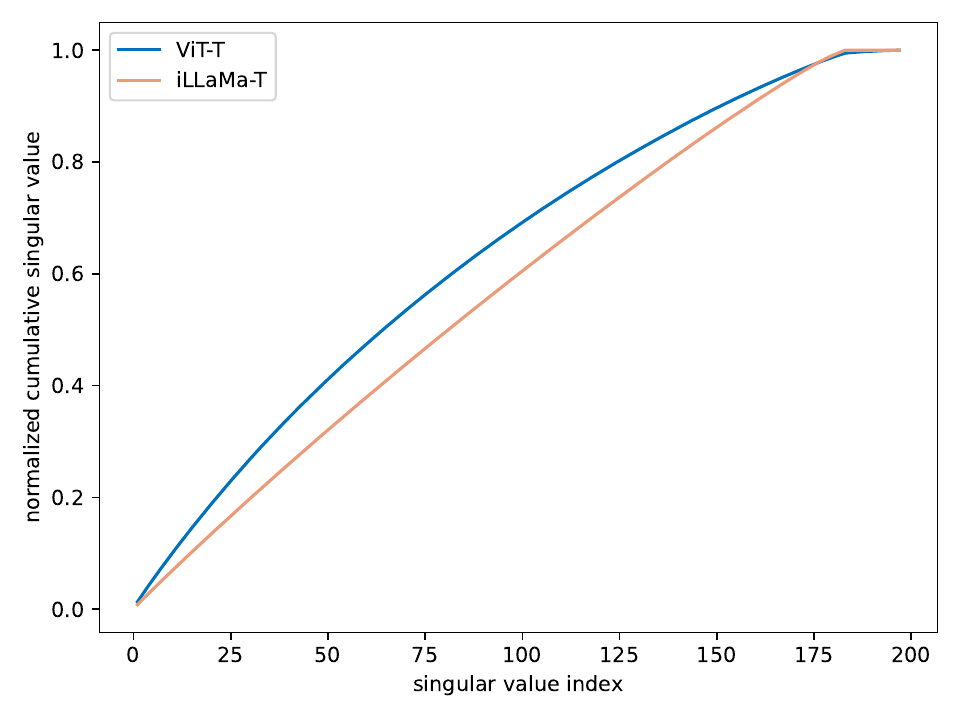}
        \caption{layer 4, head 1}
    \end{subfigure}
    \hspace{0.02\textwidth}
    \begin{subfigure}{0.3\textwidth}
        \includegraphics[width=\linewidth]{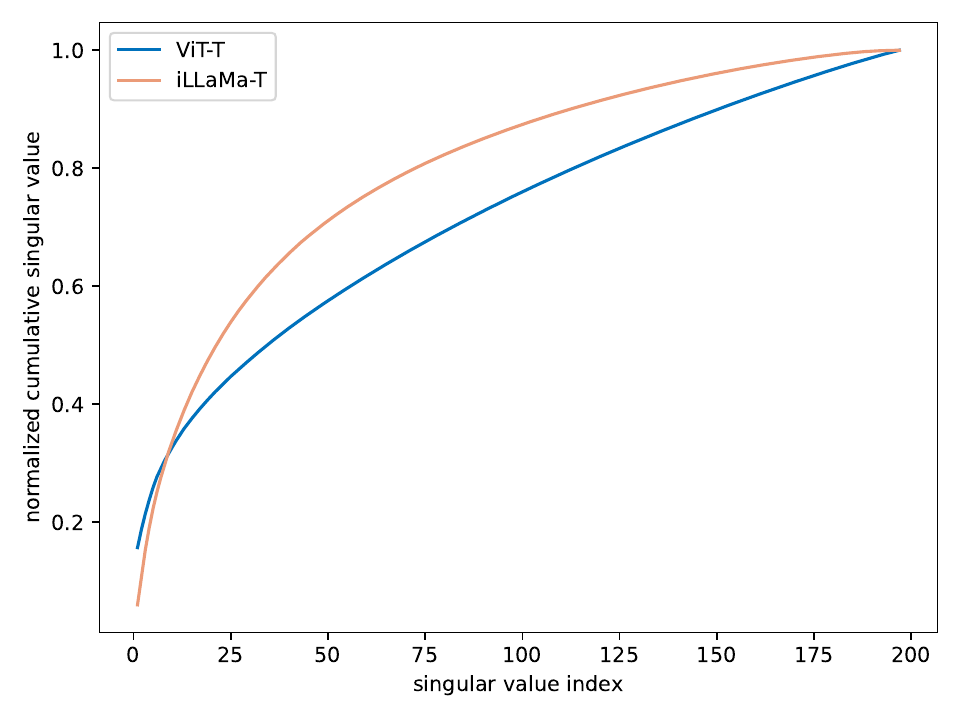}
        \caption{layer 4, head 2}
    \end{subfigure}
    \hspace{0.02\textwidth}
    \begin{subfigure}{0.3\textwidth}
        \includegraphics[width=\linewidth]{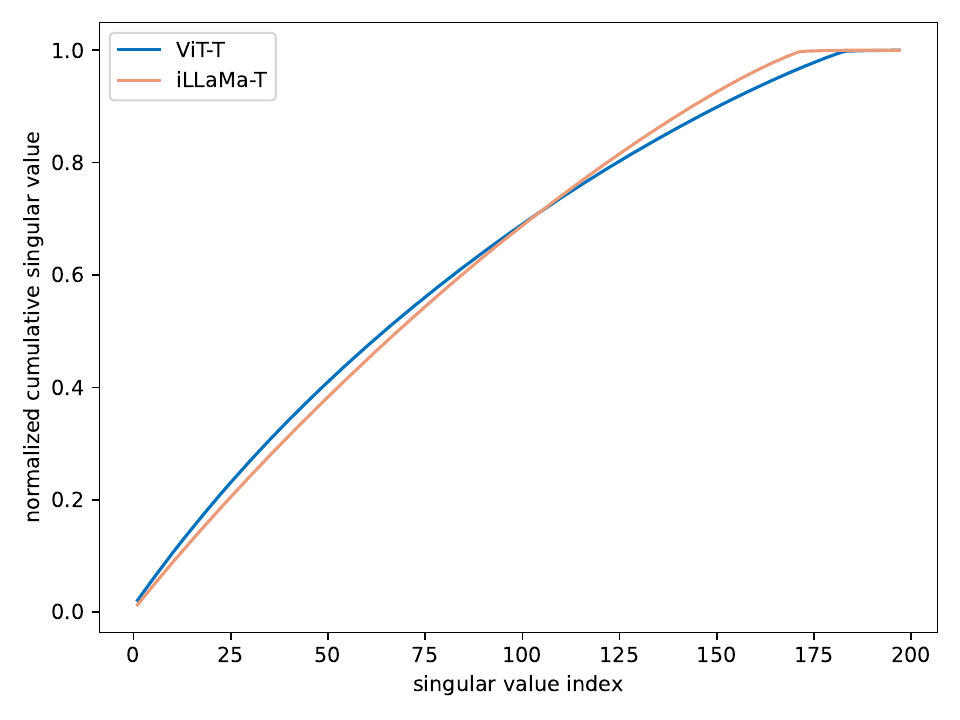}
        \caption{layer 4, head 3}
    \end{subfigure}

    \vspace{0.5cm}

    \begin{subfigure}{0.3\textwidth}
        \includegraphics[width=\linewidth]{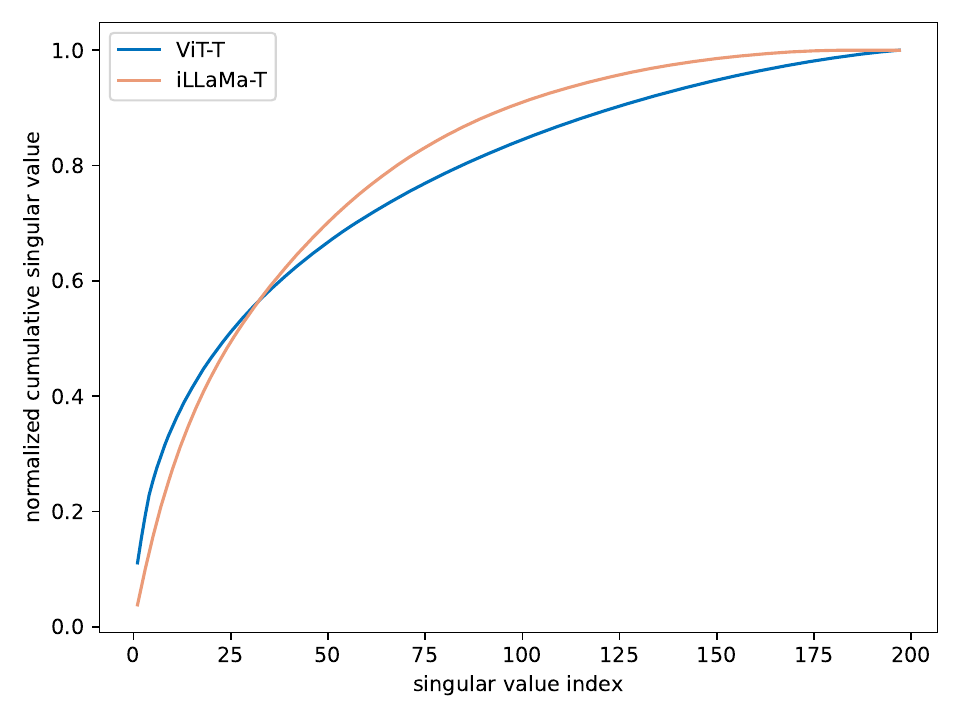}
        \caption{layer 8, head 1}
    \end{subfigure}
    \hspace{0.02\textwidth}
    \begin{subfigure}{0.3\textwidth}
        \includegraphics[width=\linewidth]{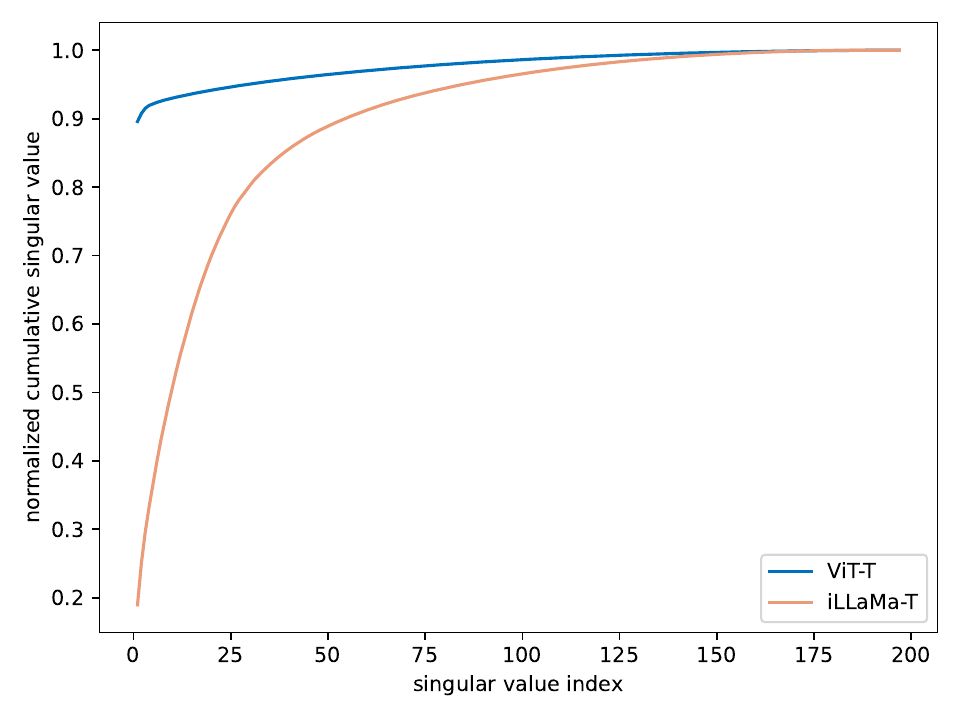}
        \caption{layer 8, head 2}
    \end{subfigure}
    \hspace{0.02\textwidth}
    \begin{subfigure}{0.3\textwidth}
        \includegraphics[width=\linewidth]{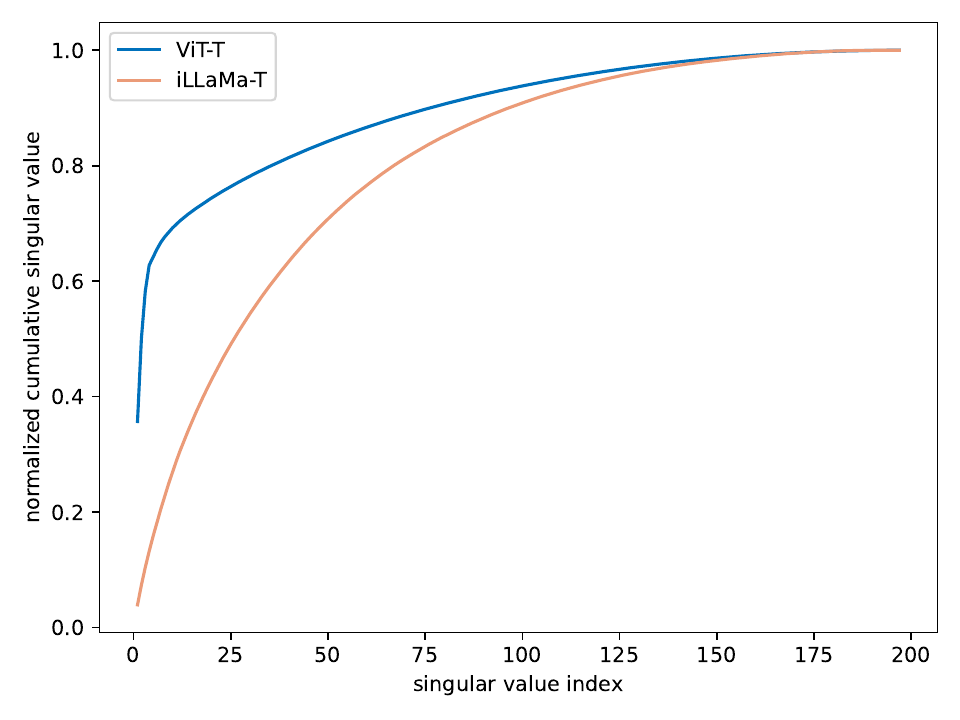}
        \caption{layer 8, head 3}
    \end{subfigure}

    \vspace{0.5cm}

    \begin{subfigure}{0.3\textwidth}
        \includegraphics[width=\linewidth]{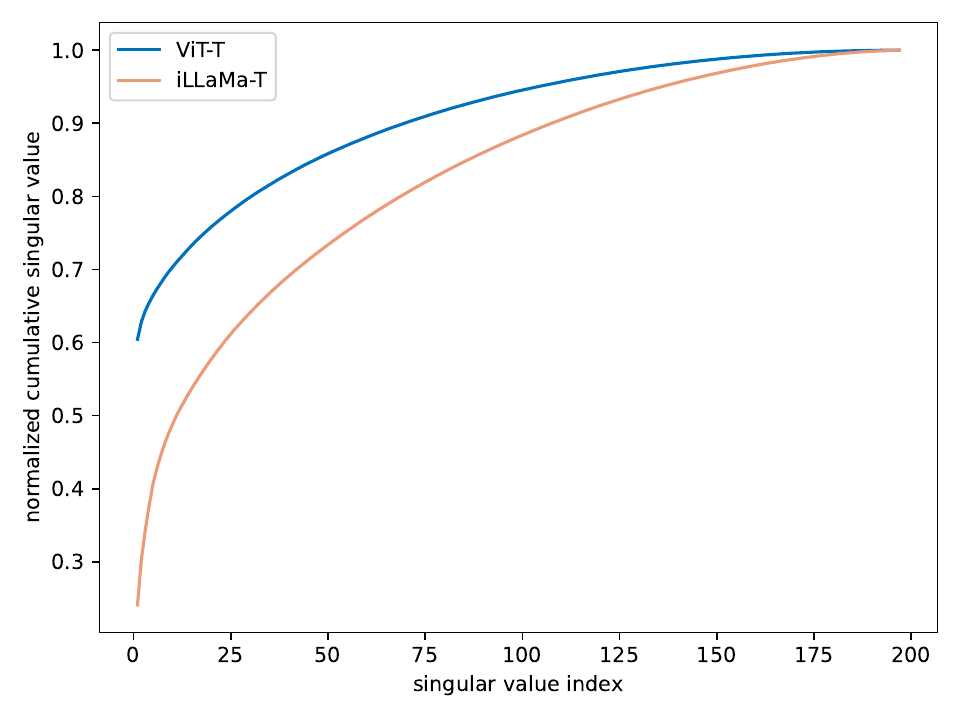}
        \caption{layer 12, head 1}
    \end{subfigure}
    \hspace{0.02\textwidth}
    \begin{subfigure}{0.3\textwidth}
        \includegraphics[width=\linewidth]{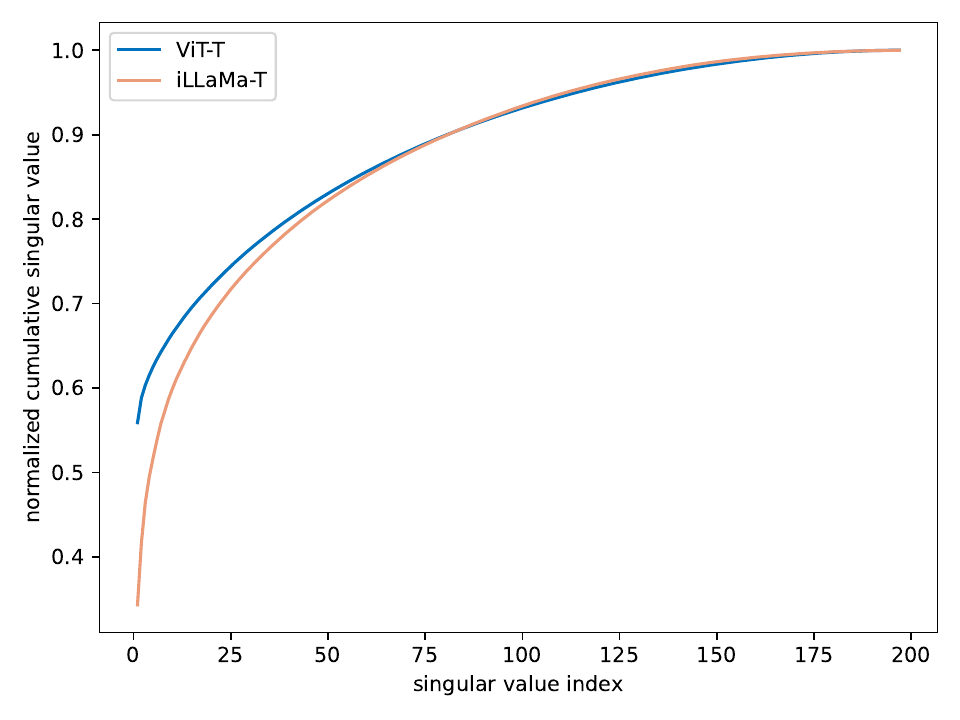}
        \caption{layer 12, head 2}
    \end{subfigure}
    \hspace{0.02\textwidth}
    \begin{subfigure}{0.3\textwidth}
        \includegraphics[width=\linewidth]{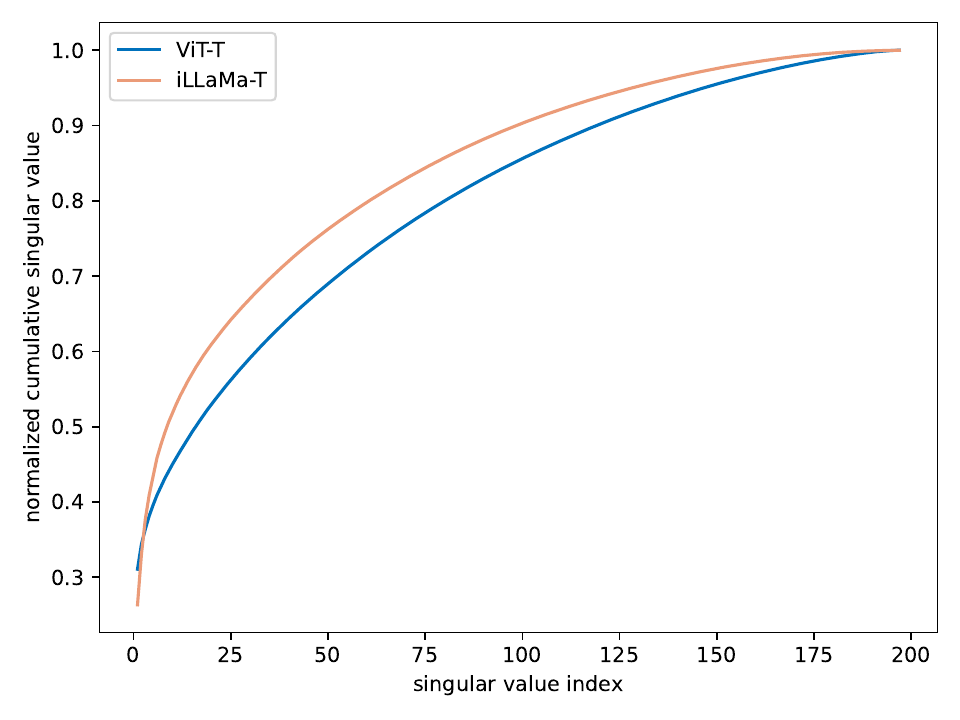}
        \caption{layer 12, head 3}
    \end{subfigure}

    \caption{Rank analysis of the self-attention matrix of all 3 heads in layer 1, 4, 8, 12 of the pretrained ViT-T and iLLaMA-T with $N=197$. In most cases, especially in shallow layers, the singular values of iLLaMa show a more uniform distribution than ViT.}
\label{fig:rank12}
\end{figure*}

\paragraph{Detailed visualization results.} 
We provide rank analysis results of all 3 heads in layer 1, 4, 8, 12 of ViT-T/16 and iLLaMA-T/16 in Figure~\ref{fig:rank12}. Besides the observation in Section~\ref{sec:3.7}, We make four observations: 
1) Not each head in each layer of iLLaMA's self-attention shows stronger concavity, suggesting that not every attention matrix of iLLaMA has a higher rank than its ViT counterpart. 
2) In most cases, particularly in the shallow layers, the distribution of singular values in iLLaMA appears more uniform than in ViT.
3) In certain attention maps (\eg, layer 8, head 2, and layer 8, head 3), the rank of ViT's attention matrix is low, resulting in an skewed distribution of information. In contrast, such extreme cases were not observed in our iLLaMA. 
4) The distribution of singular values in ViT varies significantly across different layers and heads (e.g., layer 1, head 1; layer 4, head 1; layer 8, head 1; layer 8, head 2), whereas iLLaMA's distribution appears relatively more stable. 

\section{Analysis for Soft Mask Method}
\label{sec:8.6}
In this section, we plot the training results for iLLaMA-B/16 with and without the use of the soft mask technique in Figure~\ref{fig:train_loss_base}. We can observe that the results display a similar pattern to those of iLLaMA-T/16 (Figure~\ref{fig:softmask2}(b)). 

We set the cutoff epochs to 50 and used a constant schedule. When the soft mask ends, there is a sudden increase in training loss and a steep decline in model accuracy. However, the final accuracy surpasses the baseline, and the training loss is also optimized to a lower value. Such phenomenon shows the versatility of the soft mask technique across models of varying capacities, and shows that causal self-attention can still effectively model even when a portion of the attention map is masked.

\section{Model Calibration}
\label{sec:8.7}

\begin{figure*}[h]
\centering
\begin{tabular}{ccc}
\includegraphics[width=0.3\linewidth]{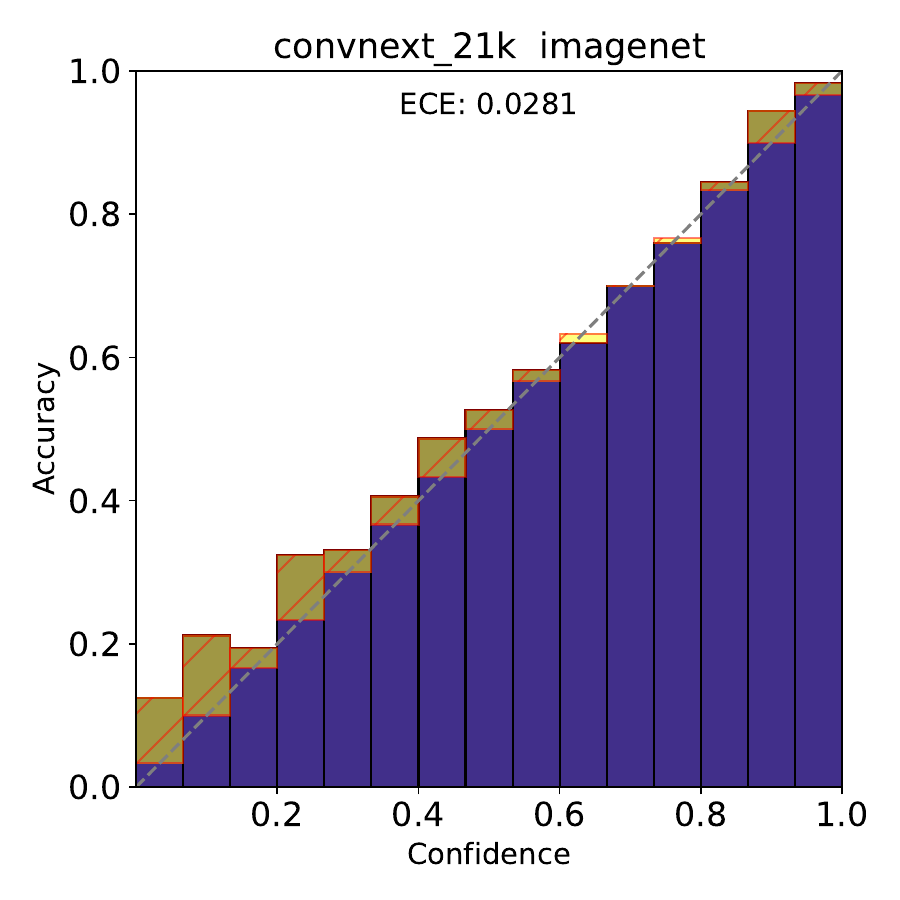} &~~~
\includegraphics[width=0.3\linewidth]{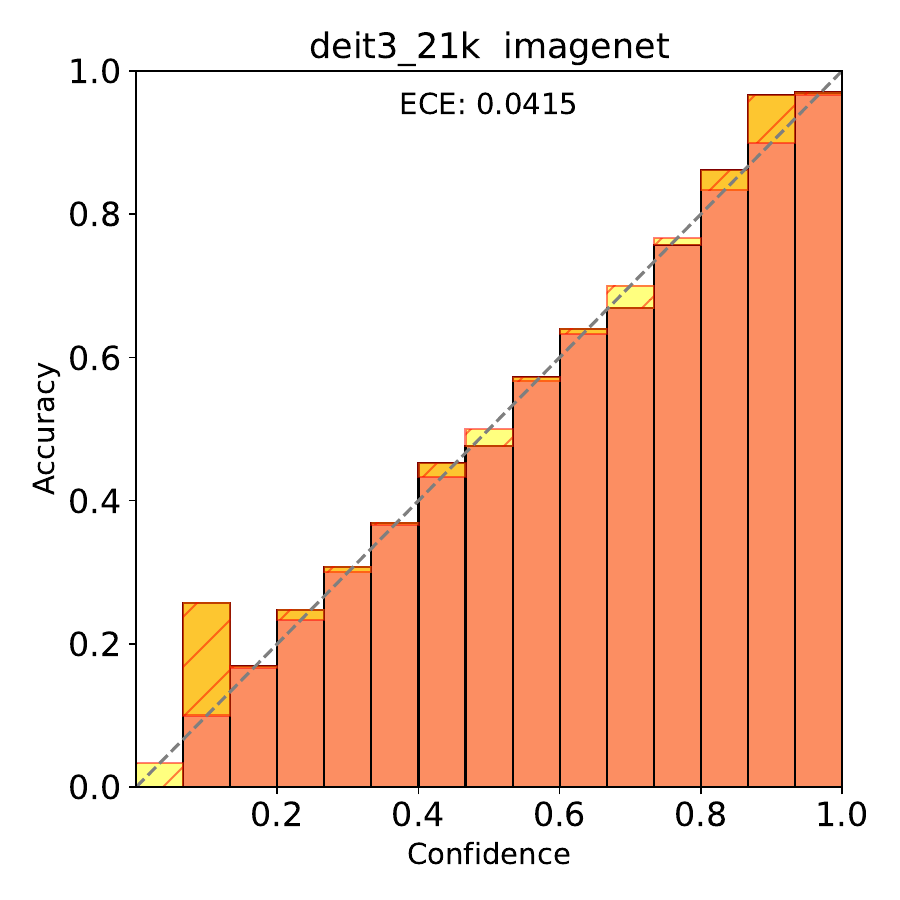}  &~~~  
\includegraphics[width=0.3\linewidth]{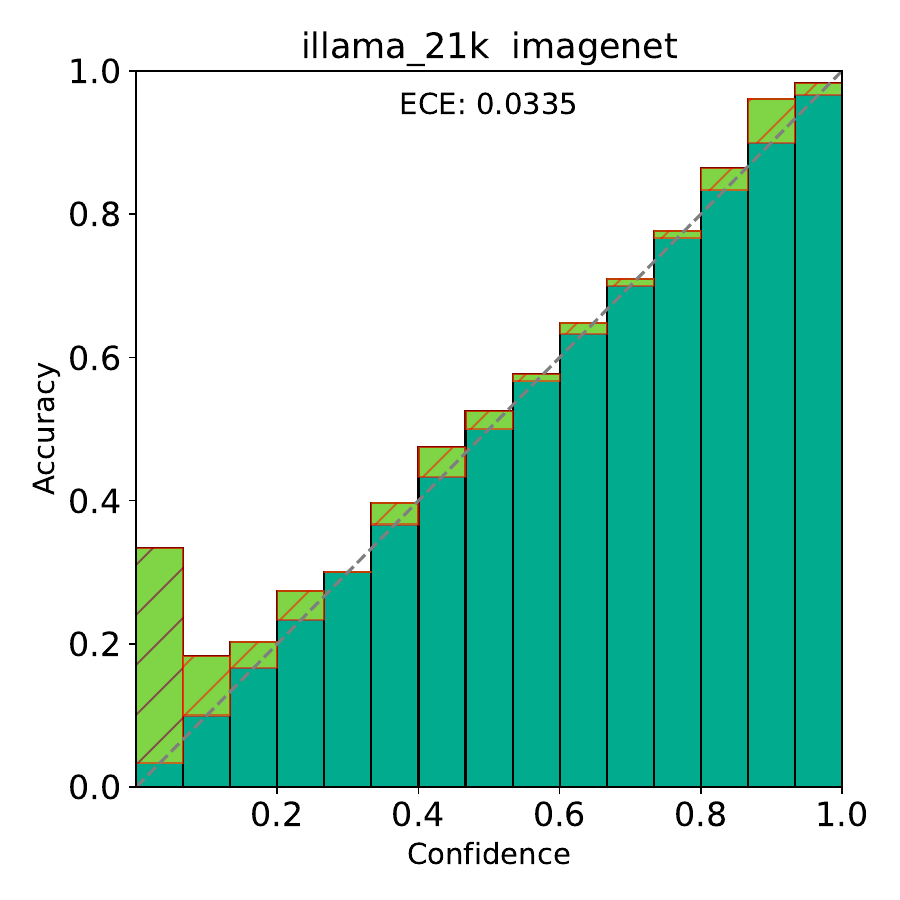}
\\
~~~(a) ConvNeXt-B & ~~~~(b) DeiT3-B & ~~~~(c) iLLaMA-B \\
\end{tabular}
\caption{Calibration results of (a) ConvNeXt-B (b) DeiT3-B and (c) iLLaMA-B pretrained on ImageNet-21K and fine-tuned on ImageNet-1K. 
}
\label{fig:calibration}
\end{figure*}

To qualitatively present the calibration property, we plot the reliability diagrams of ConvNeXt-B, DeiT3-B and the proposed iLLaMA-B using ImageNet-1K in Figure~\ref{fig:calibration}, following~\cite{vishniakov2023convnet}. For well-calibrated models, the direction of accuracy in their reliability diagrams show a roughly diagonal pattern, \ie, the difference between accuracy and confidence is small. Intuitively, the confidence of the early bins of the iLLaMA presents results below the accuracy level, indicating that iLLaMA tends to be underconfident. This observation, akin to that observed in the DeiT3, may be a common characteristic of Transformer-based architectures and was also noted in~\cite{vishniakov2023convnet}.

\section{Shape-Texture Bias}
\label{sec:8.8}
\begin{figure*}[h]
	\centering
	\includegraphics[width=0.58\linewidth]{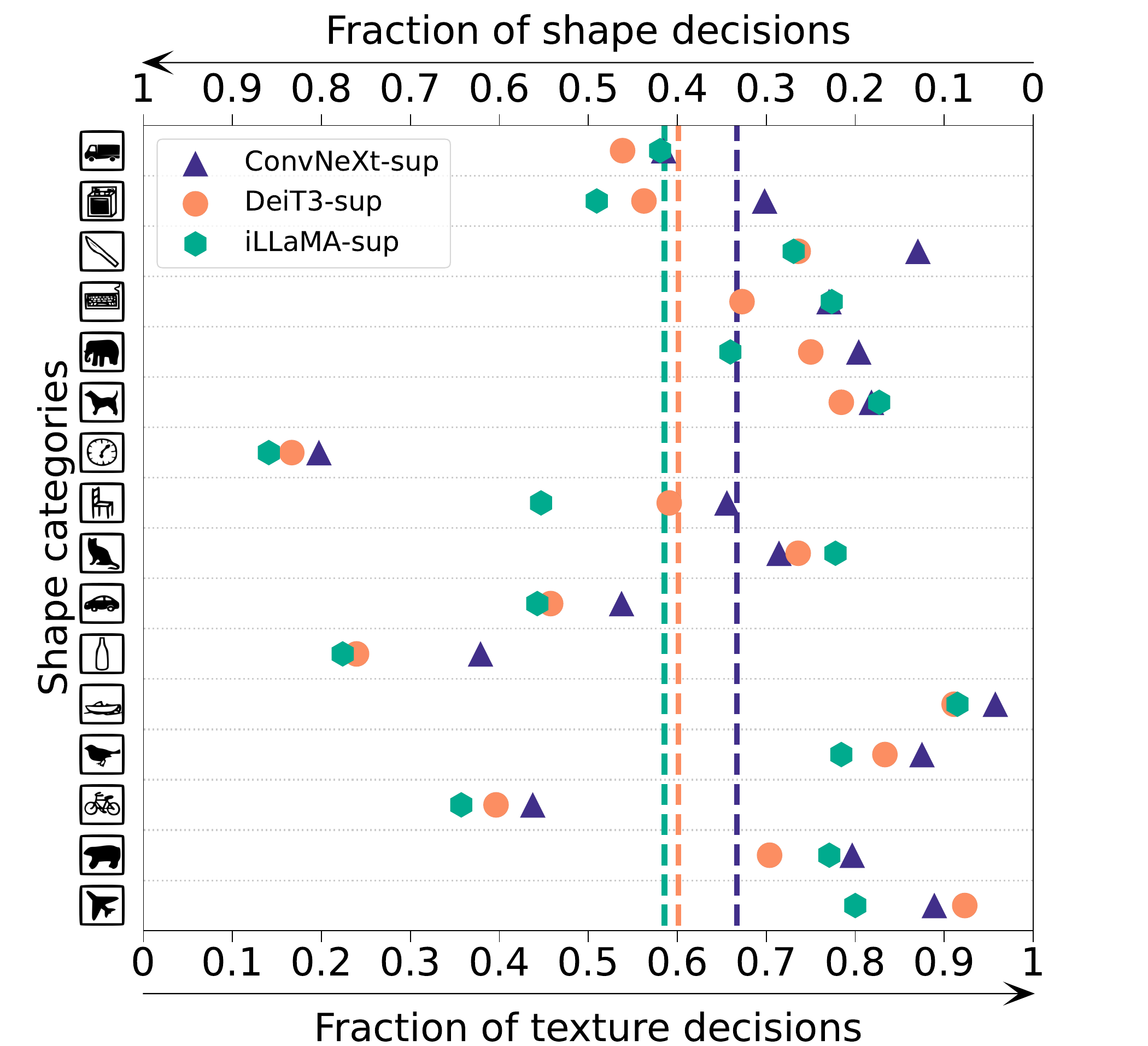}
	\caption{Shape-texture bias results of ConvNeXt-B, DeiT3-B and iLLaMA-B pre-trained on ImageNet-21K and fine-tuned on ImageNet-1K. sup: supervised learning paradigm. }
	\label{fig:shape_texture_bias}
\end{figure*}

We visualize the shape-texture bias results on cue-conflict images of ConvNeXt-B, DeiT3-B and the proposed iLLaMA-B in Figure~\ref{fig:shape_texture_bias}, following~\cite{vishniakov2023convnet}. The three dashed lines of different colors represent the average shape decision of the three models over all categories. Generally, a more leftward average shape ratio indicates that the model relies more on global shape information for recognition tasks. iLLaMA shows higher shape scores relative to ConvNeXt and DeiT3.

\section{Initializating iLLaMA Using Pre-trained LLaMA}
\label{sec:8.9}

Previous studies~\cite{zhang2024multimodal} have demonstrated that data unrelated to the image modality can be used to improve the performance of visual models. In fact, the pre-training dataset of LLaMA, which is entirely text, is irrelevant to the visual tasks that iLLaMA addresses. More important, the architectural components of iLLaMA are aligned with those of LLaMA. This alignment facilitates our exploration of using LLaMA's parameters to initialize iLLaMA, allowing us to fully exploit the potential of the weights of pre-trained LLMs. 

\begin{wraptable}{r}{0.55\textwidth}
\vspace{-1.3em}
\centering
\small
\caption{Ablation results of weight selection of iLLaMA using LLaMA2-7B pre-trained weights.}
\label{tab:weight_selection}
\begin{tabular}{lccccc}
\toprule[1.5pt]
Model & Initialization & Tiny & Small & Base \\
\midrule
iLLaMA & w/ weight selection & 74.5 & 79.9 & 81.4 \\
\gr iLLaMA & w/o weight selection & 75.0 & 79.9 & 81.6 \\
\bottomrule[1pt]
\end{tabular}
\vspace{0.0em}
\end{wraptable}
We use the pre-trained LLaMA2-7B~\cite{touvron2023llama2} to initialize our iLLaMA, instead of training from scratch. To match the dimensions of the weights, we employ the weight selection~\cite{xu2023initializing} method to initialize iLLaMA-T/S/B using a subset of the LLaMA2-7B pre-trained weights. Next, we train and evaluate the iLLaMA models, which are initialized using LLaMA2-7B, on the ImageNet-1K dataset. Other hyperparameter settings are consistent with Section~\ref{sec:4.2}. The results are shown in Table~\ref{tab:weight_selection}. 
Using LLaMA2 to initialize iLLaMA does not yield significant performance improvements. We attribute this to two main reasons:
1) The size difference between the two models is considerably large (LLaMA-2-7B's 7B parameters vs. iLLaMA-T's 5.7M parameters), resulting in a limited proportion of selected weights compared to meaningful pre-trained weights. 
2) The training strategy was not adequately optimized. We believe that when using parameter inheritance, the corresponding training strategy should also be adjusted. However, we continued to use the training recipe designed for training from scratch.

\begin{figure*}[h]
	\centering
	\includegraphics[width=0.5\linewidth]{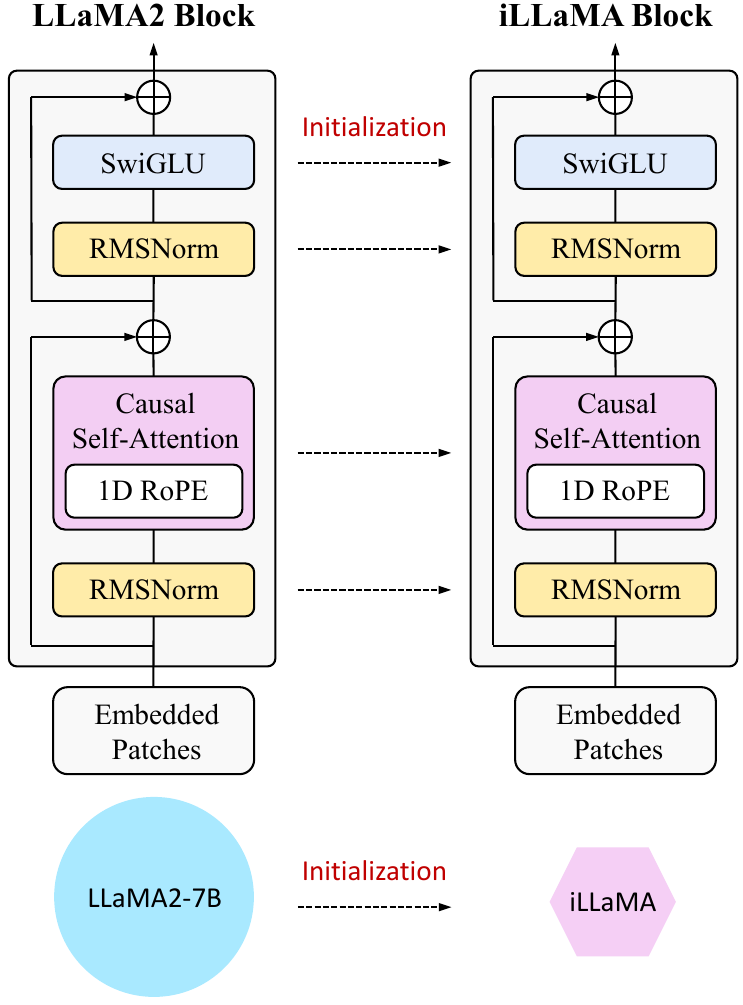}
	\caption{iLLaMA initialization by pre-trained LLaMA2-7B~\cite{touvron2023llama2} using weight selection~\cite{xu2023initializing}.}
	\label{fig:weight_selection}
\end{figure*}

\section{Loss Landscape}
\label{sec:8.10}
\begin{figure*}[h]
\centering
\begin{tabular}{ccc}
\includegraphics[width=0.4\linewidth]{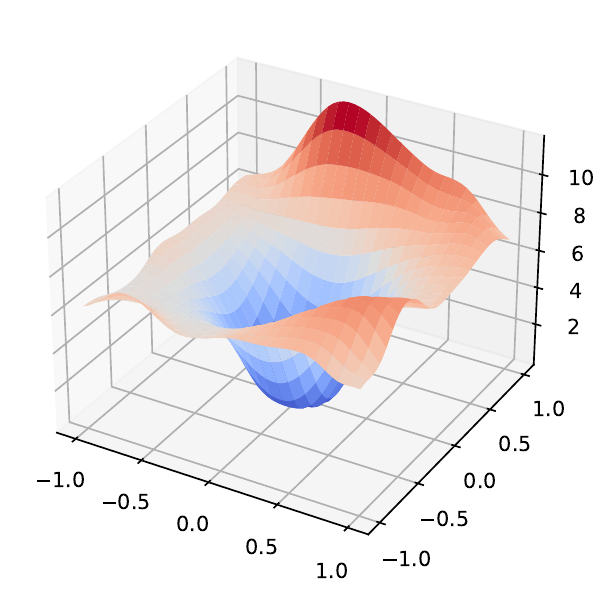} &~~~~
\includegraphics[width=0.4\linewidth]{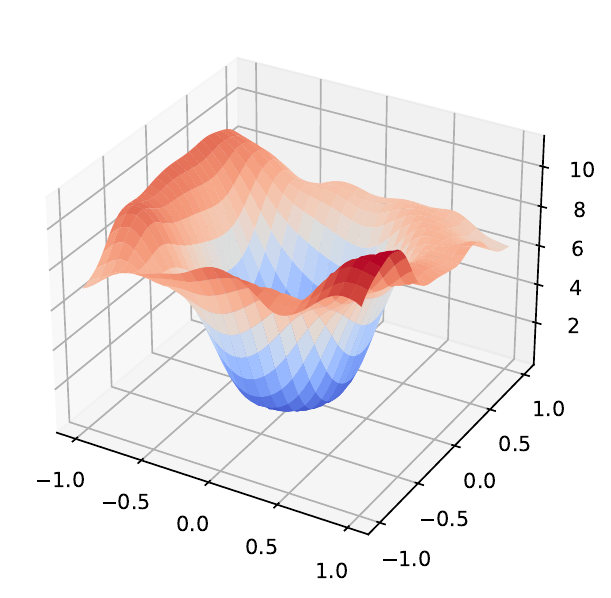}  \\
~~~(a) ViT-T/16 & ~~~~~(b) iLLaMA-T/16 \\
\end{tabular}
\caption{Loss landscape illustration of (a) ViT-T/16 and (b) iLLaMA-T/16. 
}
\label{fig:loss_landscape}
\end{figure*}
As shown in Figure~\ref{fig:loss_landscape}, we visualized the loss landscape~\cite{li2018visualizing} of the iLLaMA-T/16 and ViT-T/16. The loss landscape of ViT and iLLaMA exhibits similar patterns, with the steepness and bumps observed in ViT seeming to be softened.

\section{Limitations}
\label{sec:8.11}
We have shown that the LLaMA architecture, enhanced by the developed post-sequence [cls] and soft mask techniques, is adept at adapting to tasks such as visual recognition and semantic segmentation. However, iLLaMA's application remains predominantly within the realm of perception. In fact, such decoder-only architecture, favored by LLMs in the NLP field, can do more complex tasks, such as reasoning and generation. This may be due to their massive training data and the next sentence prediction training paradigm, which is not explored by iLLaMA. Thus, a critical validation step of aligning the architectures of text and visual models would be to construct a multi-modal large language model that fully leverages LLaMA components. In this envisioned model, both visual and textual feature extractors would be realized through the LLaMA architecture. 
Futhermore, we strongly argue that iLLaMA's successful attempts at basic supervised training strategies and classification tasks provide a foundation for more complex tasks, such as object detection and depth estimation. This represents a compelling avenue for future research. 

\section{Societal Impact}
\label{sec:8.12}
After the ChatGPT milestone in 2022, open-source architectures like LLaMA began to shine in the text domain. In the real world, images and text are the two main mediums of information and data types. For neural networks, having a unified architecture for language and vision models allows people to process these two distinct types of information using the same structure, which aids in the specialization of hardware implementation. This paper transfers the architecture widely used in language models to vision models, facilitating the achievement of this goal. The pretrained models and code provided in this paper can be used in a plug-and-play manner to serve this objective.

\end{document}